\documentclass[journal,twoside,web]{ieeecolor}
\usepackage{tmi}
\usepackage{cite}
\usepackage{amsmath,amssymb,amsfonts}
\usepackage{algorithmic}
\usepackage{graphicx}
\usepackage{textcomp}

\usepackage{booktabs}
\usepackage{multirow, makecell}
\usepackage[ruled,noend,linesnumbered]{algorithm2e}

\def\BibTeX{{\rm B\kern-.05em{\sc i\kern-.025em b}\kern-.08em
    T\kern-.1667em\lower.7ex\hbox{E}\kern-.125emX}}
\begin{document}
\title{Improving Learning of New Diseases through Knowledge-Enhanced Initialization for Federated Adapter Tuning}

\author{Danni Peng, \IEEEmembership{Member, IEEE}, Yuan Wang, \IEEEmembership{Member, IEEE}, Kangning Cai, Peiyan Ning, Jiming Xu, Yong Liu, \IEEEmembership{Senior Member, IEEE}, Rick Siow Mong Goh, \IEEEmembership{Senior Member, IEEE}, Qingsong Wei, \IEEEmembership{Senior Member, IEEE}, and Huazhu Fu, \IEEEmembership{Senior Member, IEEE}
\thanks{Manuscript submitted 2 Dec 2024. This Research is supported by the RIE2025 Industry Alignment Fund – Industry Collaboration Project (IAF-ICP) (Award No: I2301E0020) and Japan-Singapore Joint Call: Japan Science and Technology Agency (JST) and Agency for Science, Technology and Research (A*STAR) 2024 (Award No: R24I6IR141), administered by A*STAR \textit{(Corresponding author: Qingsong Wei (wei\_qingsong@ihpc.a-star.edu.sg))}.}
\thanks{Danni Peng, Yuan Wang, Yong Liu, Rick Siow Mong Goh, Qingsong Wei, and Huazhu Fu are with the Institute of High Performance Computing (IHPC), Agency for Science, Technology and Research (A*STAR), 1 Fusionopolis Way, \#16-16 Connexis, North Tower, Singapore 138632 (e-mail: dannip@ihpc.a-star.edu.sg; wang\_yuan@ihpc.a-star.edu.sg; liuyong@ihpc.a-star.edu.sg; gohsm@ihpc.a-star.edu.sg; wei\_qingsong@ihpc.a-star.edu.sg; hzfu@ieee.org).}
\thanks{Kangning Cai, Peiyan Ning, and Jiming Xu are with EVYD Technology (e-mail: kangning.cai@evydtech.com; peiyan.ning@evydtech.com; jiming.xu@evydtech.com).}
}

\maketitle

\begin{abstract}
In healthcare, federated learning (FL) is a widely adopted framework that enables privacy-preserving collaboration among medical institutions. With large foundation models (FMs) demonstrating impressive capabilities, using FMs in FL through cost-efficient adapter tuning has become a popular approach. Given the rapidly evolving healthcare environment, it is crucial for individual clients to quickly adapt to new tasks or diseases by tuning adapters while drawing upon past experiences. In this work, we introduce Federated Knowledge-Enhanced Initialization (FedKEI), a novel framework that leverages cross-client and cross-task transfer from past knowledge to generate informed initializations for learning new tasks with adapters. FedKEI begins with a global clustering process at the server to generalize knowledge across tasks, followed by the optimization of aggregation weights across clusters (inter-cluster weights) and within each cluster (intra-cluster weights) to personalize knowledge transfer for each new task. To facilitate more effective learning of the inter- and intra-cluster weights, we adopt a bi-level optimization scheme that collaboratively learns the global intra-cluster weights across clients and optimizes the local inter-cluster weights toward each client's task objective. Extensive experiments on three benchmark datasets of different modalities, including dermatology, chest X-rays, and retinal OCT, demonstrate FedKEI’s 
advantage in adapting to new diseases compared to state-of-the-art methods.
\end{abstract}


\section{Introduction}
\label{sec:intro}

\IEEEPARstart{F}{ederated} learning (FL) has gained traction in healthcare by enabling collaborative model training across institutions without sharing sensitive data \cite{rieke2020future}. With large foundation models (FMs) demonstrating strong performance across various tasks \cite{dosovitskiy2020image,radford2021learning}, integrating FMs into FL presents new opportunities for medical imaging \cite{zhuang2023foundation}. A common approach involves fine-tuning pre-trained FMs for downstream tasks in FL \cite{woisetschlager2024survey,bian2025survey}. To mitigate the substantial computation and communication overhead of full fine-tuning, recent methods adopt federated adapter tuning \cite{li2024synergizing}, which updates and transmits only lightweight adapters between the server and clients \cite{hu2021lora}. This approach enables efficient use of FMs in FL, making it particularly suitable for resource-constrained clients such as small clinics \cite{lu2023fedclip,zhao2023fedprompt}.\

In dynamic healthcare environments, FL clients often face previously unseen diseases, such as rare conditions or emerging outbreaks like COVID-19 \cite{youssef2022rapid_ai,derakhshani2022lifelonger}. To remain effective, they must continually adapt to new diagnostic tasks while leveraging prior knowledge. Federated continual learning (FCL) supports this by enabling clients to learn from evolving local task streams without sharing sensitive data \cite{yoon2021federated}. Traditional FCL methods focus on mitigating forgetting within a single shared model, under the constraint that assigning a separate model per task would incur large storage costs \cite{shoham2019overcoming,usmanova2021distillation}. However, this constraint is eased in federated adapter tuning, which fine-tunes lightweight adapters on a fixed pre-trained FM \cite{hu2021lora,li2024synergizing}. Since adapters are small (e.g.,$<$1\% of ViT with LoRA), assigning one per task is feasible, shifting the focus from “unforgetting” to knowledge transfer for better new task adaptation \cite{rusu2016progressive}. We adopt this setup by assigning a unique adapter per task and addressing the underexplored challenge of improving adaptation to new tasks.

\begin{figure}
\centerline{\includegraphics[width=0.97\linewidth]{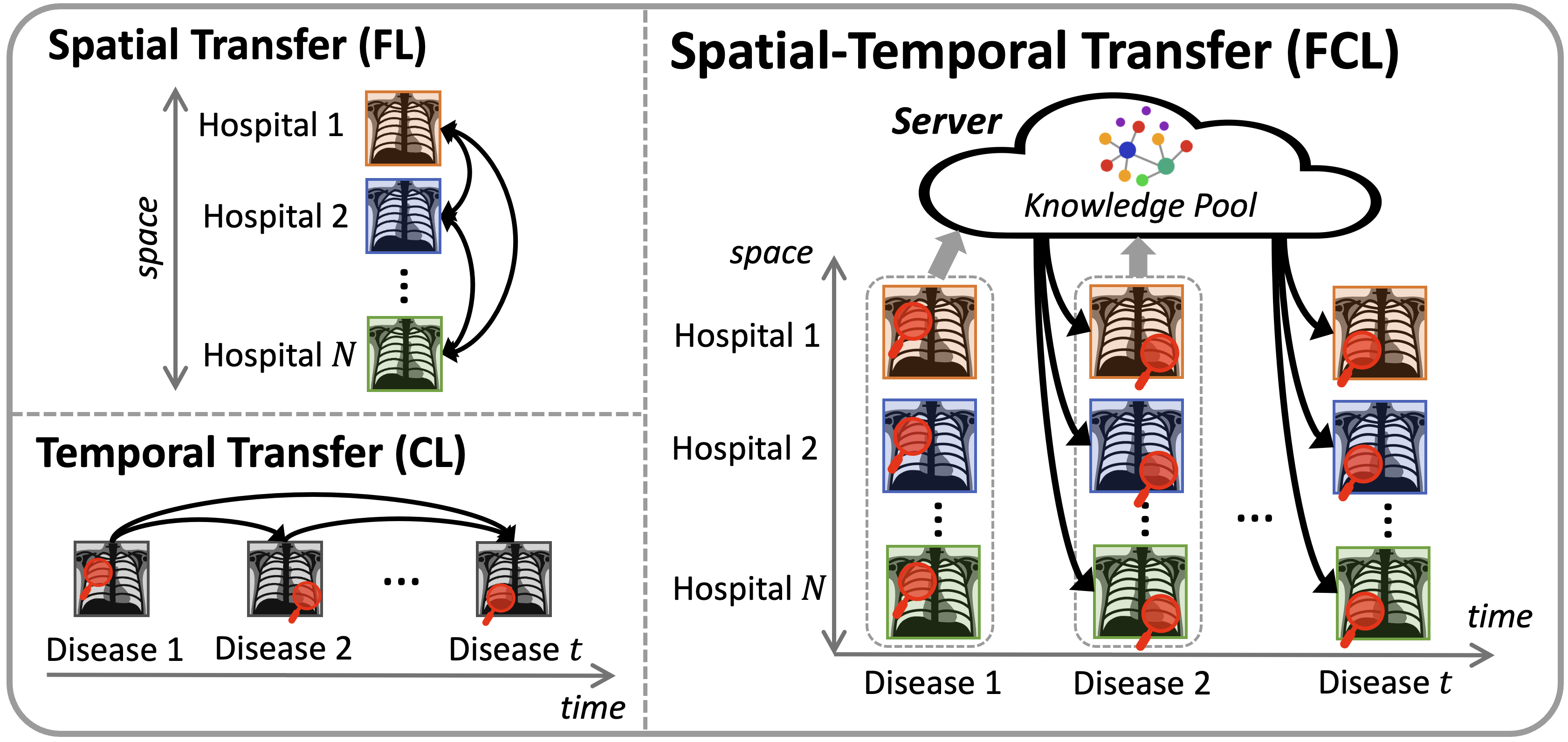}}
\caption{Illustration of spatial-temporal knowledge transfer in FCL. Left: traditional FL enables spatial transfer across clients (top), and CL enables temporal transfer across tasks (bottom). Right: our FCL framework unifies both by pooling task-specific modules from all clients at the server. When a new disease arises, each client adapts  more effectively by drawing on diverse knowledge from this shared pool.}
\label{fig:motivation}
\end{figure}

Transferring knowledge from related medical conditions is crucial for quickly adapting to new diseases. For example, insights from SARS can inform COVID-19 treatment due to their shared coronavirus origin \cite{hoffmann2020sars,rockx2020comparative}. Yet, individual hospitals often have limited, heterogeneous data. FCL enables both temporal transfer (from past tasks) and spatial transfer (from other clients), broadening each client’s knowledge base. As shown in Figure \ref{fig:motivation}, traditional FL (top left) transfers knowledge spatially across clients, and CL (bottom left) transfers knowledge temporally across tasks. Our FCL approach (right) combines both: at each step, clients upload task-specific modules (e.g., adapters) to a shared knowledge pool. When encountering a new disease, a client retrieves relevant knowledge—including its own and others’ past experiences—for better adaptation. For example, when diagnosing Atelectasis, a hospital lacking sufficient prior data may benefit from others’ clearer or more diverse cases of related conditions (e.g., Effusion or Pneumothorax), improving learning via shared visual patterns like lung collapse or opacity \cite{wang2017chestx}.\

In this work, we aim to explore the potential of leveraging spatial-temporal transfer for learning new diseases with FM adapter tuning. To this end, we propose a novel FCL framework, termed \textbf{Fed}erated \textbf{K}nowledge-\textbf{E}nhanced \textbf{I}nitialization (FedKEI), which selectively transfers valuable knowledge across clients and tasks to construct informed initializations for the adapters to learn new tasks. Specifically, the adapter and head tuned locally for each task (referred to as the task-specific modules) are sent to server, where a pool of task-specific modules is accumulated over time. Our FedKEI performs two key steps to effectively extract and transfer knowledge from the stored task-specific modules to improve new task learning: (1) global clustering of task-specific modules to generalize knowledge across tasks and clients, and (2) learning the aggregation weights across clusters (i.e., inter-cluster weights) and within each cluster (i.e., intra-cluster weights)—collectively referred to as the bi-level weights—to aggregate the task-specific modules. The resulting aggregated modules are then used to initialize the adapter and head for fine-tuning on new tasks. By this means, knowledge from previous tasks is selectively transferred and integrated into informed initializations to enhance new task learning.\

To facilitate effective learning of the bi-level aggregation weights, we adopt a bi-level optimization scheme where intra-cluster weights are learned at the server to enhance inter-cluster weight learning at the client. Inspired by meta-learning \cite{finn2017model}, this approach ensures that the cluster-specific modules, formed using shared intra-cluster weights, are more informative and capable of supporting the learning of local inter-cluster weights, leading to more effective initializations.

Overall, we summarize our main contributions as follows:
\begin{itemize}
        \item We propose a novel FCL framework, FedKEI, for improving adaptation to new diseases. The framework focuses on integrating the spatial-temporal knowledge transfer into a more informed initialization for FM adapter tuning.
        \item In our framework, we employ global clustering and bi-level aggregation learning to achieve effective knowledge transfer, where the former generalizes knowledge across different tasks and clients, and the latter selectively personalizes knowledge transfer for each task.
        \item To facilitate more effective bi-level aggregation weight learning, we introduce a novel bi-level optimization scheme to learn the global intra-cluster weights such that the subsequent learning of local inter-cluster weights is enhanced for generating effective initializations.
        \item We extensively evaluate FedKEI on three large-scale FCL datasets across different modalities and show that it achieves better performance in adapting to new diseases compared to state-of-the-art methods.
\end{itemize}

\section{Related Work}
\label{sec:rel_work}
\subsection{Task Adaptation in Federated Learning}
FL has recently gained popularity for supporting privacy-preserving collaboration among clients \cite{rieke2020future,li2021survey}. Traditional FL algorithms, such as FedAvg \cite{mcmahan2017communication}, FedProx \cite{li2020federated}, and SCAFFOLD \cite{karimireddy2020scaffold}, focus on learning a global model that performs well across all clients. While FedProx and SCAFFOLD introduce mechanisms to mitigate data heterogeneity—such as proximal regularization and control variates—they still follow a one-model-fits-all paradigm. This paradigm often falls short in scenarios with severe client heterogeneity and dynamic task distributions. Personalized federated learning (PFL) addresses this by developing personalized models tailored to each client's local objective, including meta-learning–based approaches, which learn a global initialization to facilitate client-side adaptation \cite{fallah2020personalized}, and personalized aggregation approaches, which learn client-specific aggregation weights optimized towards local objective \cite{huang2021personalized,luo2022adapt}. However, most existing PFL methods assume static client objectives and struggle with evolving tasks or shifting data, limiting their effectiveness in dynamic settings requiring continual knowledge transfer.\

FCL is a recent approach focused on continual task learning at clients, primarily aiming to mitigate forgetting. FCL methods can be broadly categorized into three classes. First, regularization-based approaches retain knowledge of previous tasks by constraining model updates, through explicit regularization terms or via knowledge distillation \cite{shoham2019overcoming,usmanova2021distillation}. Second, replay–based methods store raw samples from previous tasks or generate pseudo-examples to be used alongside new task data during training \cite{dong2022federated}. Third, architecture-based approaches assign isolated model parameters to different tasks to preserve past knowledge \cite{yoon2021federated}. Recently, \cite{tang2025afcl} explicitly identified the issue of spatial-temporal catastrophic forgetting in FCL and addressed it with a gradient-free approach. Despite recent developments, most FCL methods mainly focus on preserving performance on past tasks and preventing forgetting, with limited exploration of how knowledge from previous tasks can be harnessed to improve adaptation to new ones.\

Another related branch is federated domain generalization (FDG) \cite{li2023federated}, which aims to develop federated models that generalize well to unseen domains or tasks by learning domain-invariant features from clients with distribution shifts. ELCFS \cite{liu2021feddg} focuses on client-side learning by encouraging unbiased local training through amplitude spectrum transfer across clients, enriching local distributions for more effective domain-invariant feature extraction. In contrast, FedGA \cite{zhang2023federated} improves generalizability through server-side aggregation, adjusting client weights based on their generalization gaps. Caldarola et al. \cite{caldarola2022improving} proposes a hybrid strategy, applying sharpness-aware minimization for local training at the client side and stochastic weight averaging for model aggregation at the server. While effective, these methods address only inter-client domain shifts, assuming static local distributions. Moreover, FDG develops a generalized global model without personalization for heterogeneous local tasks. In contrast, our framework transfers knowledge across both clients and time, tailored specifically to improve learning for individual tasks.

\subsection{FM Adapter Tuning in Federated Learning}

With the rise of powerful FMs \cite{dosovitskiy2020image,radford2021learning}, there has been growing interest in integrating FMs into FL \cite{zhuang2023foundation}. Instead of full fine-tuning, tuning only the lightweight adapters provides a cost-efficient way for leveraging large FMs in FL, incurring only minimal client computation and communication \cite{hu2021lora}.

Among the earliest works, FedCLIP \cite{lu2023fedclip} demonstrates that adapter tuning outperforms full fine-tuning in FL by better retaining the rich priors of pre-trained CLIP \cite{radford2021learning}, benefiting data-scarce local tasks. Building on this, FACMIC \cite{wu2024facmic} applies adapter tuning of CLIP in the medical domain, incorporating a domain adaptation loss to mitigate client distribution shifts. In contrast, FLoRA \cite{wang2024flora} and FFA-LoRA \cite{sunimproving} focus on integrating and aggregating the existing LoRA adapters on pre-trained LLMs. FLoRA introduces a stacking-based aggregation strategy, while FFA-LoRA proposes to fine-tune only the zero-initialized matrix of LoRA to address noise during convergence. While these studies mainly concern with better implementation of adapter tuning in standard FL, our work focuses on the continual setting, leveraging knowledge transfer to improve adaptation to new tasks.

\subsection{Federated and Continual Learning in Medical Imaging}
FL has gained traction in medical imaging, with various strategies to address data heterogeneity and client-side adaptation. Feng et al. \cite{feng2022specificity} propose a shared encoder with client-specific decoders for MR image reconstruction. Xu et al. \cite{xu2022closing} introduce an ensemble framework combining global and personalized models with a model selector to handle client shift. Li et al. \cite{li2020multi} apply domain adaptation with noise-augmented fMRI data and a domain discriminator to reduce inter-client distribution gaps. ELCFS \cite{liu2021feddg}, a federated domain generalization method designed for medical image segmentation, employs a boundary-oriented episodic learning scheme to simulate domain shifts at training. \

CL has also been explored in healthcare. Wu et al. \cite{wu2023continual} mitigate forgetting in class-incremental nuclei segmentation through regularization. Amrollahi et al. \cite{amrollahi2022leveraging} enable center-incremental sepsis prediction using a hybrid of weight regularization and representation replay. Zheng et al. \cite{zheng2024towards} propose asynchronous federated continual learning with reinforcement learning and selective experience replay for modality-incremental landmark localization in 3D imaging.\

Despite recent advances, few works unify FL and CL to enable spatial-temporal knowledge transfer across clients and tasks in the medical domain. Additionally, most efforts focus on mitigating forgetting, with limited emphasis on improving new task adaptation, especially for FM adapter tuning.\

\begin{figure*}
\centerline{\includegraphics[width=0.96\linewidth]{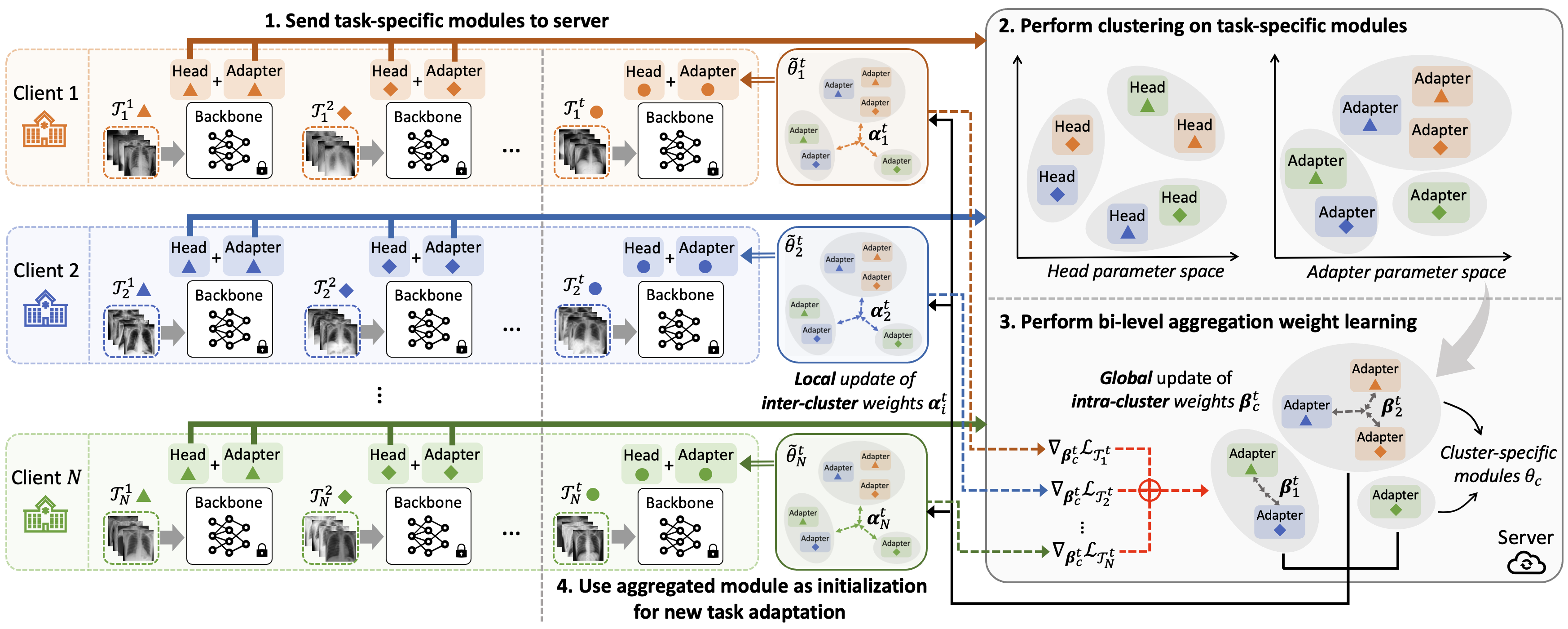}}
\caption{An overview of FedKEI. After local fine-tuning on each new task, the task-specific modules (i.e., the adapter and head) are sent to the server, where global clustering is applied to all stored modules. When new tasks arrive, the server and clients collaboratively learn bi-level aggregation weights—the local inter-cluster weights $\boldsymbol{\alpha}_i^t$ and global intra-cluster weights $\boldsymbol{\beta}_c^t$—to combine the stored task-specific modules, optimizing towards new tasks' objectives through a bi-level optimization process. Specifically, at each outer-loop step, cluster-specific modules $\theta_c$ are sent to clients, which perform inner-loop updates of $\boldsymbol{\alpha}_i^t$ by optimizing on local tasks. Gradients w.r.t. $\theta_c$ are then back-propagated through the inner-loop and returned to the server to update $\boldsymbol{\beta}_c^t$ (for simplicity, we illustrate this process using the adapter, though the same applies to the head). The final aggregated module $\tilde{\theta}_i^t$, computed using the learned bi-level weights, is used to initialize learning for the new task.}
\label{fig:overview}
\end{figure*}

\section{Methodology}
\label{sec:method}
\subsection{Preliminaries}
\label{sec:prelim}
\subsubsection{Problem Setup}
A standard FL setup involves $N$ clients and a central server, where each client $i \in [N]$ owns a local dataset $\mathcal{D}_i$. In real-world scenarios, the local data distribution at each client is not static, i.e., each client $i$ continuously encounters a local stream of tasks $\{\mathcal{T}_i^1,\mathcal{T}_i^2,\cdots\}$, where each task $\mathcal{T}_i^t \in \mathcal{D}_i$ is a subset of $\mathcal{D}_i$. To study the problem of new disease learning, we define each task to have a label set (i.e., disease classes) that differs from all the previously seen tasks of the same client, i.e., $\mathcal{Y}_i^t \neq \mathcal{Y}_i^j, \forall j<t$, where $\mathcal{Y}_i^t$ denotes the label set of $\mathcal{T}_i^t$ of client $i$. To reflect realistic medical scenarios such as emerging diseases, we assume that clients observe largely similar tasks at a given time $t$.\

In this work, we focus on improving adaptation to new tasks by leveraging knowledge from past tasks. As sharing clients’ local data is prohibited in FL, we instead utilize the task-specific models learned from previous tasks to guide current learning. Formally, at time $t$, given a new task $\mathcal{T}_i^t$ for each client $i$, we aim to learn task-specific models $\{\theta_i^t\}_{i=1}^N$ by utilizing the set of prior task-specific models $\Theta^{t-1} = \{\theta_i^j \mid i \in [N], j \in [t-1]\}$ obtained from all clients. The learning objective is defined as follows:
\begin{equation}
	\textstyle\min_{\{\theta_1^t,\cdots,\theta_N^t\}}\sum_{i=1}^N\mathcal{L}_{\mathcal{T}_i^t}(\theta_i^t;\Theta^{t-1}),
	\label{eqn:obj}
\end{equation}
where $\mathcal{L}_{\mathcal{T}_i^t}$ denotes a classification loss (e.g., cross-entropy) on the current task $\mathcal{T}_i^t$, augmented by knowledge distilled or transferred from relevant past models in $\Theta^{t-1}$.

\subsubsection{Federated Adapter Tuning for New Task Adaptation}
Adopting adapter tuning in FL offers a computation- and communication-efficient way for clients to harness the power of FMs when adapting to new tasks. Let $F^*(\cdot)$ represent a fixed, pre-trained FM backbone (e.g., ViT \cite{dosovitskiy2020image}) retained locally at each client. We attach to it an adapter module $g_{\omega}(\cdot)$, parameterized by $\omega$ (e.g., LoRA \cite{hu2021lora}). Together, the backbone and adapter form a composite feature extractor: $f_{\omega}(\cdot) = F^*(g_{\omega}(\cdot))$. Each client also maintains a classification head $h_{\phi}(\cdot)$, parameterized by $\phi$. Thus, the learnable parameters of the new task $\mathcal{T}_i^t$ consist of the adapter and the classification head: $\theta_i^t = (\omega_i^t, \phi_i^t)$. To leverage prior knowledge of the past task-specific modules $\Theta^{t-1} = \{\theta_i^j \mid i \in [N], j \in [t-1]\}$, where each $\theta_i^j = (\omega_i^j, \phi_i^j)$, the federated adapter tuning loss $\mathcal{L}_{\mathcal{T}_i^t}(\theta_i^t;\Theta^{t-1})$ for each new task $\mathcal{T}_i^t$ is defined as:
\begin{equation}
\textstyle\mathcal{L}_{\mathcal{T}_i^t}(\theta_i^t;\Theta^{t-1}) = \sum_{(x, y) \in \mathcal{T}_i^t} 
 \ell\left(h_{\phi_i^t}(f_{\omega_i^t}(x)), y|\Theta^{t-1}\right).
\label{eqn:adpt_obj}
\end{equation}

\subsection{Overview of Federated Knowledge-Enhanced Initialization (FedKEI) Framework}
To achieve the objective of improving adaptation to new tasks as described in \eqref{eqn:adpt_obj}, we propose FedKEI — a framework that leverages \textit{global clustering} and \textit{bi-level aggregation weight learning} to generate informed initializations for the adapter and head for fine-tuning on new tasks. An overview of FedKEI is shown in Figure \ref{fig:overview}. \

Generally, when a client encounters a new task, it locally fine-tunes the adapter and head (with the FM backbone fixed) and sends the task-specific modules to the server. These modules are accumulated over time into a knowledge pool, which can be leveraged for knowledge transfer to new tasks. Before local fine-tuning on a new task, the server performs two steps: (1) applies a global clustering algorithm to generalize knowledge across tasks (Section~\ref{sec:clus}); and (2) optimizes bi-level aggregation weights—global intra-cluster and personalized inter-cluster—to tailor knowledge transfer. This is done through a bi-level optimization scheme involving communication of cluster-specific modules and local task gradients between server and clients (Section~\ref{sec:bi_level}). The modules obtained by aggregating with the learned bi-level weights are used as initializations for adapter tuning on new tasks, effectively incorporating useful knowledge across time and space to facilitate new task learning (Section~\ref{sec:init}).\

In what follows, we describe each process in detail. Since our framework applies similarly to both adapters and heads—with clustering and weight learning performed separately in their respective parameter spaces—we use the term ``module" and the symbol $\theta$ to refer to both the adapter and the head throughout the description.\

\subsection{Global Clustering of Task-Specific Modules}
\label{sec:clus}
After local fine-tuning for each task at client side, the learned task-specific modules are sent to the server. Suppose we are at time $t$, the collected task-specific modules are up to time $t-1$, denoted by $\Theta^{t-1}=\{\theta_i^j|i\in[N], j\in[t-1]\}$. We apply a clustering algorithm (e.g., $k$-means++ \cite{arthur2007k}) on $\Theta^{t-1}$ to group task-specific modules with similar patterns. This serves to encourage generalization of related features among diverse tasks and facilitate knowledge transfer \cite{ghosh2020efficient,long2023multi,yao2019hierarchically}. After that, task-specific modules belonging to the same cluster are aggregated to form a set of cluster-specific modules. \

Formally, let $M = N \times (t-1)$ denote the total number of task-specific modules in $\Theta^{t-1}$, and $K$ denote the number of clusters. The cluster assignment outcome is denoted by a binary matrix $\mathbf{B} \in \{0, 1\}^{K \times M}$, where $\mathbf{B}_{c,j}$ indicates whether the $j$-th task-specific module is assigned to cluster $c$. The cluster-specific module $\theta_c$ corresponding to cluster $c \in [K]$ is obtained by aggregating the task-specific modules assigned to that cluster:
\begin{equation}
	\textstyle \theta_c = \sum_{j=1}^M \frac{\mathbf{B}_{c,j}}{\sum_{j'=1}^M \mathbf{B}_{c,j'}} \cdot \theta_j, \quad \forall c \in [K].
\label{eqn:clus}
\end{equation}

\subsection{Bi-Level Aggregation Weight Learning}
\label{sec:bi_level}
Given $K$ cluster-specific modules, each client must determine a good initialization for learning a new task. A straightforward approach is to select the best-performing cluster module \cite{ghosh2020efficient}, but this overlooks useful information from other clusters. To address this, we propose a learning-based strategy that personalizes transfer from the clustered knowledge to each new task \cite{luo2022adapt}. Specifically, we learn inter-cluster weights to combine cluster-specific modules and intra-cluster weights to refine aggregation within each cluster—together forming the bi-level aggregation weight learning. The inter-cluster weights assign importance to each cluster based on its utility for the new task, while the intra-cluster weights calibrate aggregation within each cluster to produce improved cluster-specific modules for learning personalized inter-cluster weights.\

Inspired by meta-learning, which optimizes shared components (e.g., initializations or optimizers) to improve learning across tasks \cite{finn2017model}, we adopt a bi-level optimization scheme to learn these aggregation weights. We treat the intra-cluster weights as shared components across all clients' new tasks at time $t$ and optimize them to enhance the learning of inter-cluster weights for individual tasks. Following the episodic learning paradigm of meta-learning, we optimize the global intra-cluster weights in the outer loop (at the server) and update the local inter-cluster weights in the inner loop (at the clients). We next describe the optimization procedure for both the inter- and intra-cluster weights.\

\subsubsection{Inner Updates of Local Inter-Cluster Weights} The inter-cluster weights are updated locally at the client side to directly optimize for the performance of the aggregated module on the new tasks, using the $K$ cluster-specific modules $\{\theta_c\}_{c=1}^{K}$ received from the server.\

Recalling Section \ref{sec:clus}, the cluster-specific module $\theta_c$ is obtained by aggregating the $M$ task-specific modules collected at the server with intra-cluster weights $\boldsymbol{\beta}_c=[\beta_{c,1}, \cdots, \beta_{c,M}] \in \mathbb{R}^{M}$, whose values are initialized based on the cluster assignment from the global clustering process, $\boldsymbol{\beta_c}\leftarrow \mathbf{B}_{c,:} / \sum_{j=1}^{M} \mathbf{B}_{c,j} \in \mathbb{R}^{M}$. The cluster-specific module $\theta_c$ computed using $\boldsymbol{\beta}_{c}$ is represented as:
\begin{equation}
    \theta_c(\boldsymbol{\beta}_{c})=\textstyle \sum_{j=1}^M \beta_{c,j}\cdot\theta_j, \ \ \forall c \in [K].
\label{eqn:intra_cluster}
\end{equation}
Let $\boldsymbol{\alpha}=[\alpha_1, \cdots, \alpha_K]\in\mathbb{R}^{K}$ denote the inter-cluster weights of aggregating the $K$ cluster-specific modules, initialized by $\frac{1}{K}$. The final aggregated module is obtained by:
\begin{equation}
    \textstyle\tilde{\theta}(\boldsymbol{\alpha},\boldsymbol{\beta})= \sum_{c=1}^{K}\alpha_c\cdot\theta_c(\boldsymbol{\beta}_{c}).
    \label{eqn:inter_cluster}
\end{equation}

To adapt to the new task $\mathcal{T}_i^t$ of client $i$, we update the inter-cluster weight $\boldsymbol{\alpha}$ by optimizing $\tilde{\theta}(\boldsymbol{\alpha},\boldsymbol{\beta})$ on the task objective $\mathcal{L}_{\mathcal{T}_i^t}$, while keeping $\boldsymbol{\beta}$ fixed. That is, given the set of cluster-specific modules (obtained with some fixed $\boldsymbol{\beta}$) received from the server, we perform one or several steps of updates on $\boldsymbol{\alpha}$ at the client, which we refer to as the inner-loop updates. Formally, when performing one inner-loop update, the inter-cluster weight $\boldsymbol{\alpha}^{t}_i$ for task $\mathcal{T}_i^t$ is obtained by:
\begin{equation}
\boldsymbol{\alpha}^t_i = \boldsymbol{\alpha} - \eta_1 \nabla_{\boldsymbol{\alpha}}\mathcal{L}_{\mathcal{T}_i^t}(\tilde{\theta}(\boldsymbol{\alpha},\boldsymbol{\beta})),
\label{eqn:alpha_grad}
\end{equation}
where $\eta_1$ is the inner-loop learning rate. The gradient $\nabla_{\boldsymbol{\alpha}}\mathcal{L}_{\mathcal{T}_i^t}$ is computed as $(\nabla_{\boldsymbol{\alpha}}\tilde{\theta})^{\top} \nabla_{\tilde{\theta}}\mathcal{L}_{\mathcal{T}_i^t}$, where $\nabla_{\boldsymbol{{\alpha}}}\tilde{\theta}$ is a matrix with the cluster-specific modules $[\theta_1,\cdots,\theta_K]$ as the column vectors (from \eqref{eqn:inter_cluster}). For brevity, we describe our method with a single step of inner-loop update, but the approach can easily extend to multiple updates.\

\subsubsection{Outer Updates of Global Intra-Cluster Weights} The intra-cluster weights $\boldsymbol{\beta} = [\boldsymbol{\beta}_1,\cdots,\boldsymbol{\beta}_K]$ determine how the cluster-specific modules are formed. To ensure that the cluster-specific modules are optimized for learning the new task at time $t$ across all clients, we learn the intra-cluster weights by (1) collaboratively optimizing them across all clients to enhance generalizability, and (2) ensuring that the resulting cluster-specific modules are optimal for learning the downstream inter-cluster aggregation. \

The two objectives can be achieved by optimizing the intra-cluster weights collaboratively in the outer loop such that the inter-cluster weights updated for each client in the inner loop (i.e., $\boldsymbol{\alpha}_i^t, \forall i\in[N]$) perform the best. More concretely, at time $t$, the intra-cluster weights $\boldsymbol{\beta}^t$ is optimized by encouraging better performance of the updated inter-cluster aggregated module $\tilde{\theta}(\boldsymbol{\alpha}_i^t, \boldsymbol{\beta})$ across all new tasks $\{\mathcal{T}_i^t\}_{i=1}^N$ of $N$ clients:
\begin{equation}
\textstyle\boldsymbol{\beta}^{t}=\arg\min_{\boldsymbol{\beta}}\sum_{i=1}^N\mathcal{L}_{\mathcal{T}_i^t}(\tilde{\theta}(\boldsymbol{\alpha}_i^t,\boldsymbol{\beta})).
\label{eqn:beta_obj}
\end{equation}
Considering the intra-cluster weight $\boldsymbol{\beta}_{c}$ for each cluster $c$ separately, the gradient descent update is given by:
\begin{equation}\
\textstyle\boldsymbol{\beta}_{c} \leftarrow \boldsymbol{\beta}_{c} - \eta_2 \sum_{i=1}^N\nabla_{\boldsymbol{\beta}_c}\mathcal{L}_{\mathcal{T}_i^t}(\tilde{\theta}(\boldsymbol{\alpha}_i^t,\boldsymbol{\beta})), \ \ \forall c \in [K],
\label{eqn:beta_grad}
\end{equation}
where $\eta_2$ is the outer-loop learning rate.  We can see that the gradient for updating $\boldsymbol{\beta}_c$ is the sum of the gradients of individual tasks $\nabla_{\boldsymbol{\beta}_c}\mathcal{L}_{\mathcal{T}_i^t}, \forall i \in [N]$, which by chain rule, can be expressed as:
\begin{equation}
\nabla_{\boldsymbol{\beta}_c}\mathcal{L}_{\mathcal{T}_i^t} = (\nabla_{\boldsymbol{\beta}_c}\theta_c)^{\top} (\nabla_{\theta_c}\tilde{\theta}(\boldsymbol{\alpha}_i^t,\boldsymbol{\beta}))^{\top} \nabla_{\tilde{\theta}(\boldsymbol{\alpha}_i^t,\boldsymbol{\beta})}{\mathcal{L}_{\mathcal{T}_i^t}}.
\label{eqn:beta_grad_exp_raw}
\end{equation}
Here, the first term $\nabla_{\boldsymbol{\beta}_c}\theta_c$ is simply a matrix with the task-specific modules $[\theta_1,\cdots,\theta_M]$ as the column vectors (from \eqref{eqn:intra_cluster}). The second term $\nabla_{\theta_c}\tilde{\theta}(\boldsymbol{\alpha}_i^t,\boldsymbol{\beta})$ is the derivative of the updated inter-cluster aggregated module $\tilde{\theta}(\boldsymbol{\alpha}_i^t, \boldsymbol{\beta})$ w.r.t. the cluster-specific module $\theta_c$, where $\boldsymbol{\alpha}_i^t$ is obtained by updates in the inner loop leveraging the cluster-specific modules $\{\theta_c\}_{c=1}^K$ (see \eqref{eqn:alpha_grad} and \eqref{eqn:inter_cluster}). Hence, evaluating $\nabla_{\theta_c}\tilde{\theta}(\boldsymbol{\alpha}_i^t,\boldsymbol{\beta})$ involves back-propagating through the inner loop conducted at the client side. Deriving from \eqref{eqn:alpha_grad}, it can be obtained that $\nabla_{\theta_c}\tilde{\theta}(\boldsymbol{\alpha}_i^t,\boldsymbol{\beta})=\alpha_{i,c}^t I - \eta_1\theta_c (\nabla_{\tilde{\theta}(\boldsymbol{\alpha},\boldsymbol{\beta})}\mathcal{L}_{\mathcal{T}_i^t})^\top$, where $\alpha_{i,c}^t$ is the updated inter-cluster weight associated with cluster $c$, and $\nabla_{\tilde{\theta}(\boldsymbol{\alpha},\boldsymbol{\beta})}\mathcal{L}_{\mathcal{T}_i^t}$ is the task gradient w.r.t. the inter-cluster aggregated module using the initial $\boldsymbol{\alpha}$. \

Generally, computing the outer-loop gradients in \eqref{eqn:beta_grad_exp_raw} requires (a) the task-specific modules stored at the server, which constitutes the first term $\nabla_{\boldsymbol{\beta}_c}\theta_c$, and (b) the gradients w.r.t. $\theta_c$ derived through the inner-loop $\nabla_{\theta_c}{\mathcal{L}_{\mathcal{T}_i^t}}=(\nabla_{\theta_c}\tilde{\theta}(\boldsymbol{\alpha}_i^t,\boldsymbol{\beta}))^{\top} \nabla_{\tilde{\theta}(\boldsymbol{\alpha}_i^t,\boldsymbol{\beta})}{\mathcal{L}_{\mathcal{T}_i^t}}$.
Since the inner-loop updates of $\boldsymbol{\alpha}$ is conducted at the client side, we compute $\nabla_{\theta_c}{\mathcal{L}_{\mathcal{T}_i^t}}$ at clients and send it to the server to compute the outer-loop gradient $\nabla_{\boldsymbol{\beta}_c}\mathcal{L}_{\mathcal{T}_i^t}$ in \eqref{eqn:beta_grad_exp_raw}. We then aggregate $\nabla_{\boldsymbol{\beta}_c}\mathcal{L}_{\mathcal{T}_i^t}$ across $N$ clients to perform updates on $\boldsymbol{\beta}_c$ in \eqref{eqn:beta_grad}. In our implementation, we perform only one outer-loop update. Hence, the additional computation and communication overhead introduced remains manageable. Since only gradients related to the cluster-specific modules are shared with the server, privacy is preserved.

\begin{figure}
\centerline{\includegraphics[width=\linewidth]{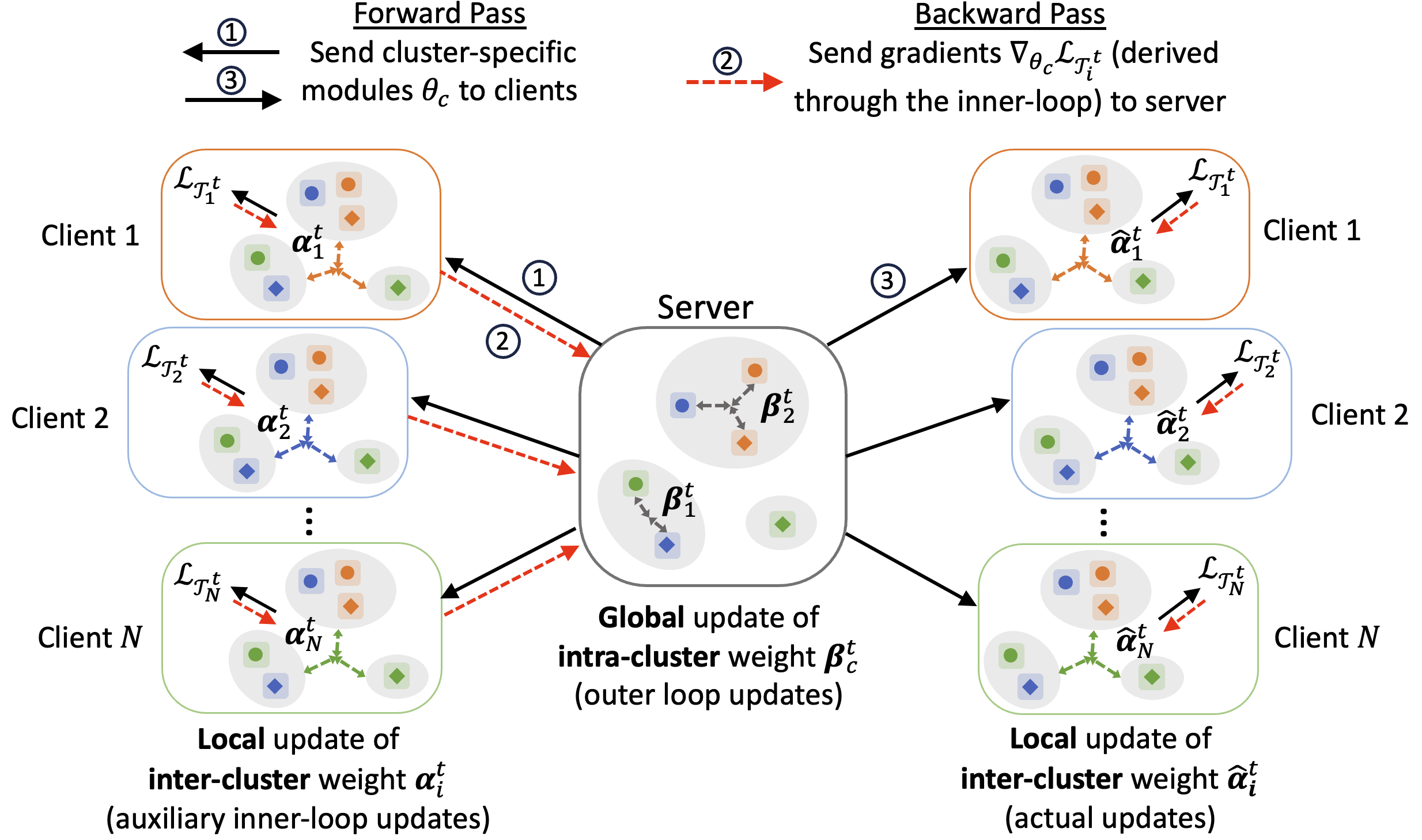}}
\caption{An illustration of bi-level aggregation weight learning. The cluster-specific modules $\theta_c$, weighted by global intra-cluster weights $\boldsymbol{\beta}_c$, are sent to clients (outer-loop forward pass). Based on the received $\theta_c$, each client locally updates inter-cluster weights $\boldsymbol{\alpha}_i^t$ by minimizing the task loss $\mathcal{L}_{\mathcal{T}_i^t}$ (inner-loop). Upon completion, the gradients $\nabla_{\theta_c}\mathcal{L}_{\mathcal{T}_i^t}$ are derived through the inner-loop updates and sent back to the server (outer-loop backward pass). After outer-loop optimization, the learned $\boldsymbol{\beta}_c^t$ is forwarded for learning the actual inter-cluster weight $\hat{\boldsymbol{\alpha}}_i^t$.}
\label{fig:fast_slow}
\end{figure}

\subsubsection{Actual Learning of Inter-Cluster Weights} Upon completion of the bi-level optimization, we obtain a set of effective cluster-specific modules with updated intra-cluster weights $\boldsymbol{\beta}^t$, generalized for all the tasks at time $t$. On the updated cluster-specific modules, each client $i$ then performs actual learning of personalized inter-cluster weight $\hat{\boldsymbol{\alpha}}_i^t$ for the new task $\mathcal{T}_i^t$:
\begin{equation}
\begin{aligned}
\textstyle\hat{\boldsymbol{\alpha}}^{t}_i=\arg\min_{\boldsymbol{\alpha}}\mathcal{L}_{\mathcal{T}_i^t}(\tilde{\theta}(\boldsymbol{\alpha},\boldsymbol{\beta}^t)).
\end{aligned}
\label{eqn:alpha_obj_updated}
\end{equation}
Note that the inner-loop updates described earlier are only auxiliary for optimizing the intra-cluster weights. Leveraging the optimized intra-cluster weights, the inter-cluster weights $\hat{\boldsymbol{\alpha}}^{t}_i$ computed here contribute to the actual initialization for learning task $\mathcal{T}_i^t$. Figure \ref{fig:fast_slow} illustrates the full process of bi-level aggregation weight learning.\

\subsection{Initialization for New Task Adaptation}
\label{sec:init}
Once we obtain the learned intra-cluster weights $\boldsymbol{\beta}^t$ and the personalized inter-cluster weights $\hat{\boldsymbol{\alpha}}_i^t$ tailored to task $\mathcal{T}_i^t$, we generate the final aggregated module $\tilde{\theta}_i^t=\tilde{\theta}(\hat{\boldsymbol{\alpha}}_i^t, \boldsymbol{\beta}^t)$ and use it as initialization for adapter tuning on new task $\mathcal{T}_i^t$:
\begin{equation}
\min_{\theta_i^t\leftarrow\tilde{\theta}_i^t}\mathcal{L}_{\mathcal{T}_i^t}(\theta_i^t).
\label{eqn:new_task_adpt}
\end{equation}
Generally, the learned bi-level weights $\hat{\boldsymbol{\alpha}}_i^t, \boldsymbol{\beta}^t$ collectively determine how to leverage knowledge from tasks stored in the knowledge pool to facilitate learning of new tasks. 
Algorithm \ref{alg:algo} summarizes the overall workflow of the FedKEI framework. 

\begin{algorithm}[tb]
\footnotesize
\SetKwInput{KwInput}{Input}                
\SetKwInput{KwOutput}{Output}              
\caption{FedKEI Framework}
\label{alg:algo}
\KwInput{Number of clients $N$; Number of tasks $M$; Client $i$'s task data $\{\mathcal{T}_i^1, \cdots, \mathcal{T}_i^M\}$, $\forall i \in [N]$; Fixed backbone $F^*$; Number of clusters $K$; Learning rates for inner- and outer-loop $\eta_1$, $\eta_2$; Number of update steps for inner- and outer-loop $\tau$, $n$.}
\KwOutput{Learned task-specific modules $\{\theta_i^t|i\in[N],t\in[M]\}$}
\For{$t\in\{1, \cdots, M\}$}
{
\textbf{\# Global clustering of task-specific modules} \\
Cluster $N \times (t-1)$ task-specific modules collected at server.\\
Compute cluster-specific modules $\{\theta_c\}_{c=1}^K$ by \eqref{eqn:clus}.\\
\textbf{\# Outer-loop updates of global intra-cluster weights} \\
\For{each outer step}
{
\For{client $i \in [N]$ in parallel}
{Download cluster-specific modules $\{\theta_c\}_{c=1}^K$ from server.\\
\textbf{\# Inner-loop updates of local inter-cluster weights} \\
\For{each inner step}
{Perform \textit{local} update of inter-cluster weight $\boldsymbol{\alpha}_i^t$ by \eqref{eqn:alpha_grad}.}
Compute $\{\nabla_{\theta_c}{\mathcal{L}_{\mathcal{T}_i^t}}\}_{c=1}^K$ and send to server.
}
Compute $\{\nabla_{\boldsymbol{\beta}_c}\mathcal{L}_{\mathcal{T}_i^t}\}_{i=1}^N$ for all clients by \eqref{eqn:beta_grad_exp_raw}.\\
Perform \textit{global} update of intra-cluster weight $\boldsymbol{\beta}^t_c$ by \eqref{eqn:beta_grad}.\\
}
Compute optimized cluster-specific modules $\{\theta_c^t\}_{c=1}^K$ with the learned $\boldsymbol{\beta}^t$.\\
\textbf{\# Actual learning of local inter-cluster weights} \\
\For{client $i \in [N]$ in parallel}
{Download optimized cluster-specific modules $\{\theta_c^t\}_{c=1}^K$ from server.\\
\While{not converged}
{Perform actual update of inter-cluster weight $\hat{\boldsymbol{\alpha}}_i^t$ by optimizing \eqref{eqn:alpha_obj_updated}.}
}
\textbf{\# Initialization for new task adaptation} \\
Use the aggregated module $\tilde{\theta}(\hat{\boldsymbol{\alpha}}_i^t, \boldsymbol{\beta}^t)$ as the initialization for adapter tuning on $\mathcal{T}_i^t$ by \eqref{eqn:new_task_adpt}.
}
\end{algorithm}

\section{Experiments}
\label{sec:exp}
\subsection{Experimental Setup}
\subsubsection{Datasets and Settings}

\begin{figure}
        \centering
        \includegraphics[width=0.8\linewidth]{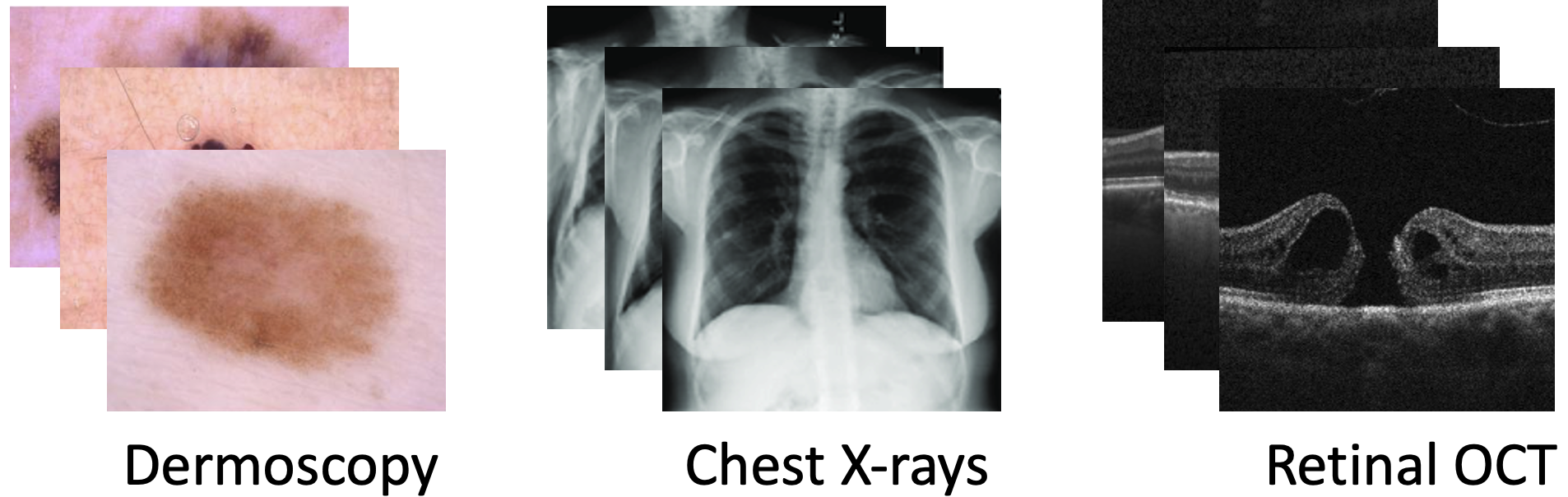} 
        \caption{Medical imaging modalities used in our experiments.}
        \label{fig:modality}
\end{figure}

We evaluate FedKEI on three FL datasets across different medical imaging modalities for disease classification: (1) skin lesion identification from dermoscopic images; (2) chest disease diagnosis from X-rays; and (3) eye disease classification from retinal OCT. These datasets cover diverse imaging techniques and conditions, allowing us to assess robustness across clinical scenarios. Figure \ref{fig:modality} shows examples of modalities used in our experiments. \

We construct tasks to simulate the FCL setting, defining a task order such that, at each time step, the task (i.e., the disease) encountered by all clients is largely the same. This synchronous task order setup reflects real-world scenarios in which medical institutions often face the same new diseases—such as emerging illnesses or pandemic outbreaks—around the same time.\

\begin{figure}
        \centering
        \includegraphics[width=0.95\linewidth]{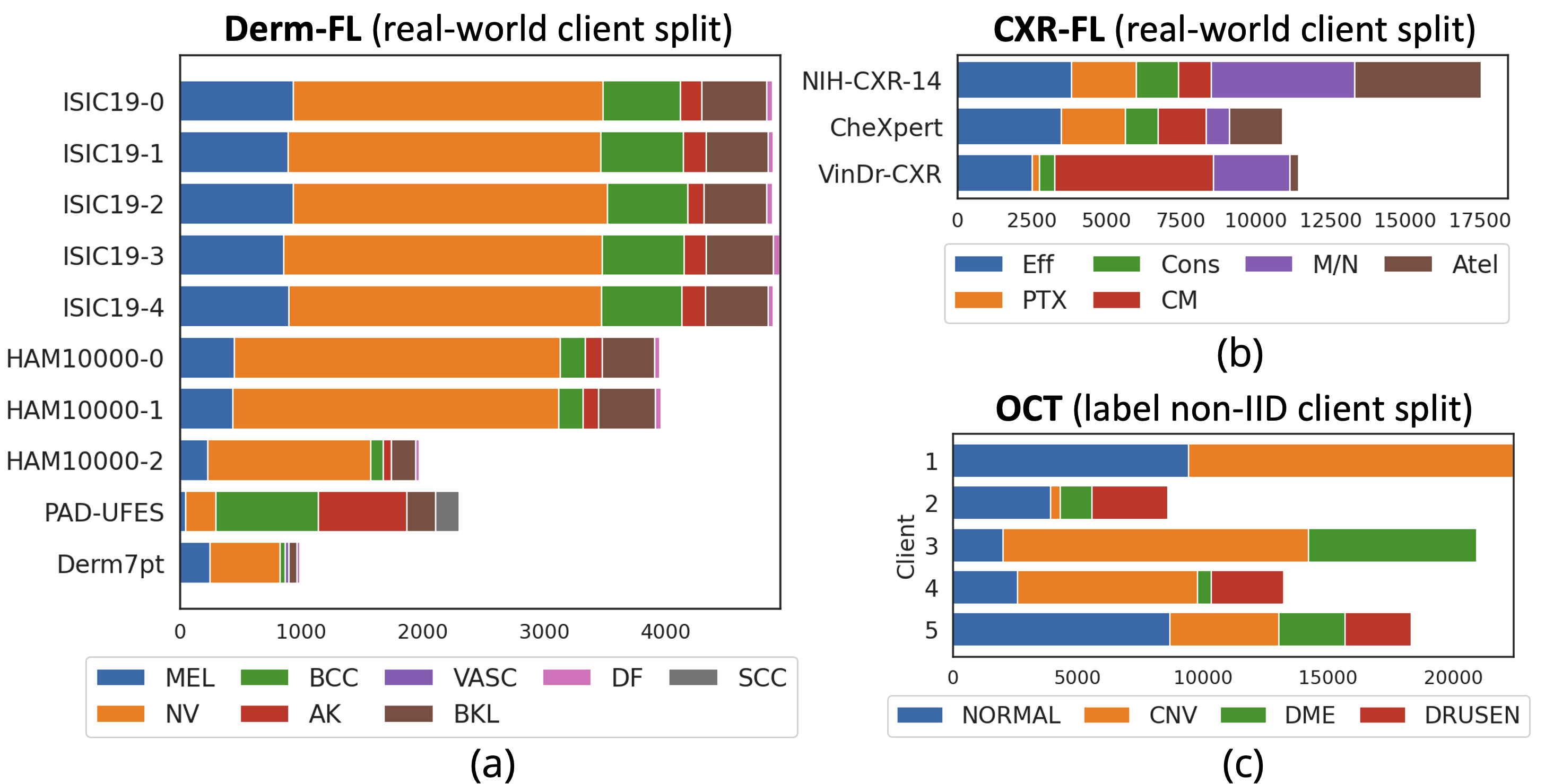}
        \caption{Data distribution among clients for (a) Derm-FL, (b) CXR-FL, and (c) OCT datasets.}
        \label{fig:client_dist}
\end{figure}

\begin{figure}
        \centering
        \includegraphics[width=0.92\linewidth]{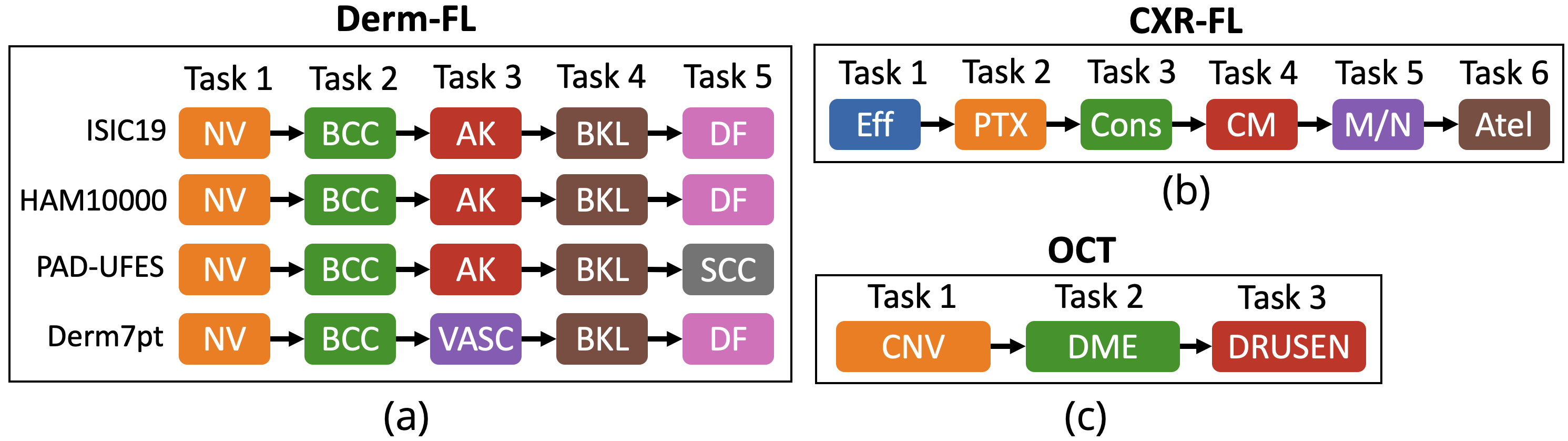}
        \caption{Order of disease learning for (a) Derm-FL, (b) CXR-FL, and (c) OCT datasets.}
        \label{fig:task_order}
\end{figure}

\textbf{Derm-FL Dataset}: Following \cite{yan2023label}, we collect data from four public skin datasets — ISIC19 \cite{codella2019skin}, HAM10000 \cite{tschandl2018ham10000}, PAD-UFES \cite{pacheco2020impact}, and Derm7pt \cite{kawahara2018seven} — and split them into ten clients to simulate an FL setting ($N=10$). Figure \ref{fig:client_dist}a shows the data distribution among ten clients. The combined dataset includes 37,607 dermoscopic images across eight skin lesion types: MEL, NV, BCC, AK, VASC, BKL, DF, and SCC. To simulate continual adaptation, we construct five sequential tasks per dataset ($M=5$), each involving identifying a new disease from those previously seen. Figure \ref{fig:task_order}a shows the order of disease learning for the four datasets. For example, the first task is to identify NV from MEL, the second BCC from \{MEL, NV\}, and so on. While tasks are mostly aligned across clients, some datasets lack certain diseases (e.g., PAD-UFES lacks DF, Derm7pt lacks AK). In such cases, we substitute available alternatives (e.g., VASC for AK in Derm7pt). This mild relaxation of task alignment reflects real-world variability in disease prevalence and data availability, allowing us to evaluate FedKEI’s robustness under such conditions.\

\textbf{CXR-FL Dataset}: We compile chest X-ray images from three public datasets—NIH-CXR-14 \cite{wang2017chestx}, CheXpert \cite{irvin2019chexpert}, and VinDr-CXR \cite{nguyen2022vindr}—treating each as a client to simulate an FL setting ($N=3$). We retain seven common classes (including `No Finding’ and six diseases: Eff, PTX, Cons, CM, M/N, and Atel), resulting in 79,010 images. Figure \ref{fig:client_dist}b shows the data distribution among three clients. By treating each disease as a task, we construct six sequential tasks ($M=6$). The first task is to identify Eff from No Finding, the second task is to identify PTX from \{No Finding, Eff\}, etc. Figure \ref{fig:task_order}b shows the task order for CXR-FL dataset. Since all three clients contain the six diseases, the same task order is applied to all clients.\

\textbf{OCT Dataset}: We also evaluate our method on the OCT dataset \cite{kermany2018large}, which contains 84,495 retinal images spanning four conditions: Normal, CNV, DME, and Drusen. To simulate label non-IID scenario, we split each class among five clients following $Dir(0.5)$ ($N=5$). Figure \ref{fig:client_dist}c shows the data distribution among five clients. We construct three tasks from the four classes ($M=3$). The first task is to identify CNV from Normal, the second is to identify DME from \{Normal, CNV\}, etc. Figure \ref{fig:task_order}c shows the task order for OCT dataset.\

\subsubsection{Implementation Details}
For fair comparisons, all methods use the same FM adapter tuning setup. For our main experiments, we adopt ViT-B/16 \cite{dosovitskiy2020image} pre-trained on ImageNet as the fixed backbone $F^*$ and fine-tune LoRA adapters \cite{hu2021lora}. For each new task, the adapter and head are fine-tuned locally for 30 epochs with a learning rate of 0.005 and batch size of 64 across all three datasets. Communication between server and clients occurs only at the start of each new task to transmit aggregated knowledge. All images are preprocessed to match the ViT-B/16 input: grayscale images (from CXR-FL and OCT) are converted to pseudo-RGB by channel replication, resized to 224×224, and normalized using ImageNet statistics, consistent with the FM’s pretraining protocol.\

For our FedKEI, we set both the inner-loop learning rate $\eta_1$ (for inter-cluster weight $\boldsymbol{\alpha}$) and the outer-loop learning rate $\eta_2$ (for intra-cluster weight $\boldsymbol{\beta}$) to 0.05. In each inner loop, we update the inter-cluster weight $\boldsymbol{\alpha}$ on local task for 1 epoch with a batch size of 64 (i.e., the number of inner-loop steps varies based on task data size). The number of outer-loop steps is set to 1. For clustering, we perform $k$-means++ \cite{arthur2007k} and tune the number of clusters $K$ for both adapters and heads in $\{3, 5, 7, 9\}$. We use SGD optimizer for all gradient updates.\

\subsubsection{Evaluation Metrics}
We evaluate new task adaptation using two metrics: (1) final AUC at the last epoch of each task, and (2) Learning Curve Area (LCA), calculated by averaging the AUC across all epochs of each task. The former assesses how well a task is learned and the latter indicates the speed of learning a task \cite{chaudhry2018efficient}. A larger LCA reflects faster learning, as it indicates that higher performance is reached earlier during training—often due to better initialization or effective knowledge transfer from prior tasks—resulting in a greater cumulative area under the learning curve. For all experiments, we conduct three trials with different seeds and report the mean and standard deviation.\

\subsection{Baseline Comparison}
We compare FedKEI with 10 baselines across four FL categories. Traditional FL methods include: (1)FedAvg \cite{mcmahan2017communication}, which averages task-specific modules from all clients for initialization; (2)FedProx \cite{li2020federated}, which adds a proximal term to improve generalization; and (3)FFA-LoRA \cite{sunimproving}, which enhances LoRA aggregation by partially freezing LoRA's weights. FCL methods prevent forgetting: (4) FedCurv \cite{shoham2019overcoming}, a regularization-based method preserving prior tasks; and (5) FLwF \cite{usmanova2021distillation}, a distillation-based method aligning current and previous task logits. FDG methods improve generalization to unseen tasks: (6)ELCFS \cite{liu2021feddg}, which shares amplitude spectra across clients; and (7)FedGA \cite{zhang2023federated}, which enhances aggregation for better global generalization. PFL methods personalize models: (8) Per-FedAvg \cite{fallah2020personalized}, a meta-learning method for fast adaptation; (9) FedAMP \cite{huang2021personalized}, which computes aggregation weights based on model similarity; and (10) APPLE \cite{luo2022adapt}, which learns aggregation weights by optimizing client objectives. Lastly, we include a naive baseline Rand, which randomly initializes the adapter and head for new task.\

\setlength{\tabcolsep}{3pt}
\begin{table}[t]
\centering
\caption{Baseline comparisons on Derm-FL dataset. We report the individual task performance (AUC) averaged over 10 clients as well as the overall mean AUC and LCA over 5 tasks.}
  		\begin{tabular}{l c c c c c c c}
    		\toprule
    		\multirow{2}[2]{*}{\textbf{Method}} & \multicolumn{5}{c}{Individual Task AUC} & \multicolumn{2}{c}{Overall}\\
    		\cmidrule(lr){2-6} \cmidrule(lr){7-8}
                & \scriptsize Task 1 & \scriptsize Task 2 & \scriptsize Task 3 & \scriptsize Task 4 & \scriptsize Task 5 & AUC & LCA \\
    		\midrule
    		Rand\textsuperscript{‡} & 87.92 & 93.33 & 87.16 & 78.38 & 71.04 & 83.57\tiny$\pm$0.12 & 81.15\tiny$\pm$0.05 \\
                FedAvg\textsuperscript{‡} & 87.92 & 93.40 & 88.94 & 78.64 & 72.89 & 84.36\tiny$\pm$0.11 & 81.50\tiny$\pm$0.10 \\
                FedProx\textsuperscript{†} & 86.54 & 91.88 & \underline{88.95} & 78.31 & 74.01 & 84.14\tiny$\pm$0.18 & \underline{82.18}\tiny$\pm$0.23 \\
                FFA-LoRA\textsuperscript{†} & 88.19 & 93.38 & 88.58 & \underline{78.71} & 73.66 & 84.50\tiny$\pm$0.35 & 81.56\tiny$\pm$0.30 \\
                FedCurv\textsuperscript{‡} & 87.92 & 93.00 & 88.43 & 77.60 & 69.39 & 83.27\tiny$\pm$0.00 & 80.32\tiny$\pm$0.00 \\
                FLwF\textsuperscript{‡} & 87.92 & 93.21 & 88.18 & 78.61 & 70.04 & 83.59\tiny$\pm$0.07 & 80.50\tiny$\pm$0.18 \\
                ELCFS\textsuperscript{‡} & 87.37 & 92.51 & 87.36 & 77.18 & 71.38 & 83.16\tiny$\pm$0.09 & 78.96\tiny$\pm$0.12 \\
                FedGA\textsuperscript{‡} & 87.92 & 93.51 & 88.91 & 78.69 & 73.81 & 84.57\tiny$\pm$0.06 & 81.79\tiny$\pm$0.06 \\
                Per-FedAvg\textsuperscript{‡} & 88.06 & 93.51 & 87.51 & 77.82 & 75.45 & 84.47\tiny$\pm$0.15 & 81.21\tiny$\pm$0.09 \\
                FedAMP\textsuperscript{‡} & 87.92 & 91.72 & 87.04 & 76.99 & 75.40 & 83.82\tiny$\pm$0.04 & 79.98\tiny$\pm$0.00 \\
                APPLE\textsuperscript{‡} & 87.92 & \underline{93.65} & 87.66 & 78.06 & \underline{75.83} & \underline{84.63}\tiny$\pm$0.16 & 81.61\tiny$\pm$0.07 \\
                \midrule
                FedKEI & 87.92 & \textbf{94.71} & \textbf{89.93} & \textbf{79.55} & \textbf{80.48} & \textbf{86.52}\tiny$\pm$0.10 & \textbf{85.07}\tiny$\pm$0.08 \\
    		\bottomrule
  		\end{tabular}
\label{tbl:baseline_derm}
\end{table}

\begin{figure*}
\centerline{\includegraphics[width=0.91\linewidth]{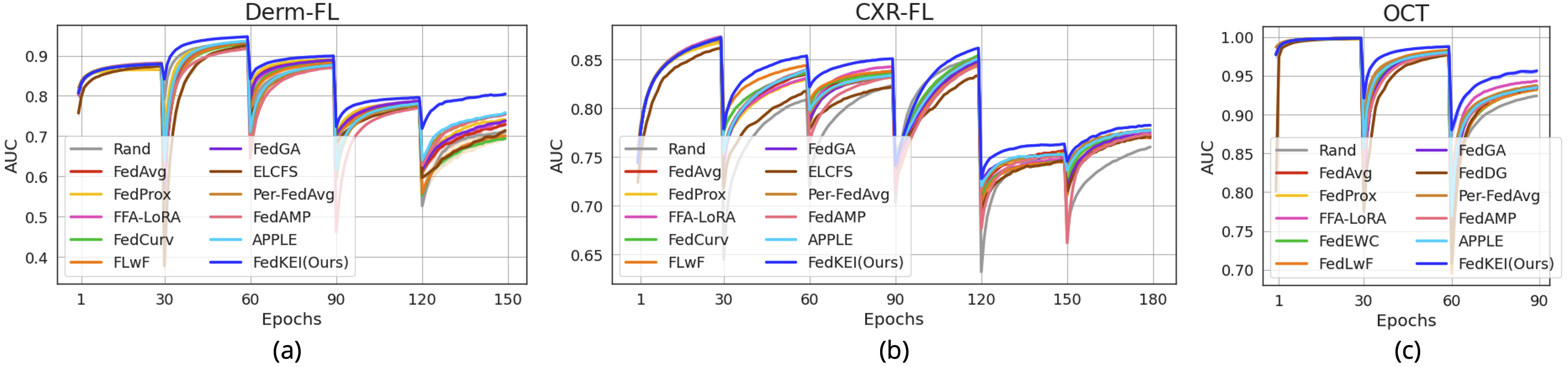}}
\caption{Learning curves for individual tasks of FedKEI and the compared baselines on (a) Derm-FL, (b) CXR-FL, and (c) OCT datasets.}
\label{fig:overall_plot}
\end{figure*}

Table \ref{tbl:baseline_derm}, \ref{tbl:baseline_cxr} and \ref{tbl:baseline_oct} present results on Derm-FL, CXR-FL, and OCT datasets, respectively, which include the final AUC of each individual task and the overall AUC and LCA averaged across all tasks. To assess the significance of FedKEI’s performance gains over the baselines, we conducted Welch’s t-test. The resulting $p$-values for AUC and LCA are denoted by superscripts following each method name in the tables: \textsuperscript{‡} for $p<0.001$, \textsuperscript{†} for $p<0.01$, and \textsuperscript{*} for $p<0.05$. For all columns, the best and second-best scores are shown in bold and underline, respectively. Note that no score is highlighted for Task 1, as all methods perform the same as Rand—except FedProx, FFA-LoRA, ELCFS, and Per-FedAvg, which involve modifications to the local fine-tuning process.\

First, from Table \ref{tbl:baseline_derm}, we observe that FedKEI consistently outperforms all baselines on the Derm-FL dataset. Specifically, it achieves improvements of 1.89\% and 2.89\% in overall AUC and LCA, respectively, over the second-best methods (APPLE, FedProx). These results suggest that FedKEI is effective in enhancing both the accuracy and efficiency of new task adaptation. Among the baselines, FCL methods (FedCurv, FLwF) perform poorly—often worse than the naive Rand—since merely preserving performance on old tasks does not guarantee effective transfer to new tasks and limits adaptability. FDG method ELCFS also underperforms, suggesting that enhancing local diversity without task-specific focus can hurt performance. Traditional FL methods (FedAvg, FedProx, FFA-LoRA) generally perform well, showing that aggregating past task-specific modules offers more useful initializations than Rand. Among PFL baselines, rule-based FedAMP lags behind, while learning-based Per-FedAvg and APPLE perform well, highlighting the benefits of direct task-specific optimization. Our FedKEI, by learning the bi-level aggregation specifically on the new tasks, generates the most effective initializations and achieves the best overall performance.\

\setlength{\tabcolsep}{1.8pt}
\begin{table}[t]
\centering
\caption{Baseline comparisons on CXR-FL dataset. We report the individual task performance (AUC) averaged over 3 clients as well as the overall mean AUC and LCA over 6 tasks.}
  		\begin{tabular}{l c c c c c c c c}
    		\toprule
    		\multirow{2}[2]{*}{\textbf{Method}} & \multicolumn{6}{c}{Individual Task AUC} & \multicolumn{2}{c}{Overall}\\
    		\cmidrule(lr){2-7} \cmidrule(lr){8-9}
                & \scriptsize Task 1 & \scriptsize Task 2 & \scriptsize Task 3 & \scriptsize Task 4 & \scriptsize Task 5 & \scriptsize Task 6 & AUC & LCA \\
    		\midrule
          	Rand\textsuperscript{*} & 87.23 & 80.88 & 82.35 & 85.28 & 75.23 & 76.03 & 81.17\tiny$\pm$0.44 & 78.40\tiny$\pm$0.54 \\
                FedAvg\textsuperscript{*} & 87.23 & 83.56 & 83.80 & 84.87 & \underline{75.68} & 76.96 & 82.02\tiny$\pm$0.02 & 79.93\tiny$\pm$0.24 \\
                FedProx\textsuperscript{*} & 86.74 & 82.99 & 83.17 & 84.36 & 75.32 & 77.19 & 81.63\tiny$\pm$0.61 & 79.84\tiny$\pm$0.60 \\
                FFA-LoRA\textsuperscript{‡} & 87.33 & 83.09 & \underline{84.24} & 85.26 & 74.78 & 77.56 & 82.04\tiny$\pm$0.01 & 79.95\tiny$\pm$0.02 \\
                FedCurv\textsuperscript{*} & 87.23 & 83.60 & 83.50 & \underline{85.42} & 74.64 & 77.14 & 81.92\tiny$\pm$0.20 & 80.13\tiny$\pm$0.24 \\
                FLwF\textsuperscript{†} & 87.23 & \underline{84.36} & 83.75 & 84.60 & 74.45 & 77.06 & 81.91\tiny$\pm$0.18 & 80.04\tiny$\pm$0.03 \\
                ELCFS\textsuperscript{*} & 86.14 & 81.77 & 82.19 & 83.38 & 74.61 & 77.10 & 80.86\tiny$\pm$0.41 & 78.63\tiny$\pm$0.37 \\
                FedGA\textsuperscript{†} & 87.23 & 83.87 & 83.23 & 84.79 & 75.06 & 77.33 & 81.92\tiny$\pm$0.15 & 79.92\tiny$\pm$0.17 
 \\
                 Per-FedAvg\textsuperscript{*} & 87.09 & 83.85 & 83.22 & 85.16 & 75.13 & \underline{77.78} & 82.04\tiny$\pm$0.23 & \underline{80.17}\tiny$\pm$0.19 \\
                FedAMP\textsuperscript{*} & 87.23 & 83.94 & 83.22 & 84.44 & 75.16 & 77.27 &  81.88\tiny$\pm$0.26 & 79.49\tiny$\pm$0.21 \\
                APPLE\textsuperscript{*} & 87.23 & 83.91 & 83.34 & 85.11 & 75.23 & 77.73 &  \underline{82.09}\tiny$\pm$0.25 & \underline{80.17}\tiny$\pm$0.21 \\
                \midrule
                FedKEI & 87.23 & \textbf{85.36} & \textbf{85.07} & \textbf{86.16} & \textbf{76.33} & \textbf{78.25} & \textbf{83.07}\tiny$\pm$0.05 & \textbf{81.33}\tiny$\pm$0.06  \\
    		\bottomrule
  		\end{tabular}
\label{tbl:baseline_cxr}
\end{table}

\setlength{\tabcolsep}{6pt}
\begin{table}[t]
\centering
\caption{Baseline comparisons on OCT dataset. We report the individual task performance (AUC) averaged over 5 clients as well as the overall mean AUC and LCA over 3 tasks.}
  		\begin{tabular}{l c c c c c}
    		\toprule
    		\multirow{2}[2]{*}{\textbf{Method}} & \multicolumn{3}{c}{Individual Task AUC} & \multicolumn{2}{c}{Overall}\\
    		\cmidrule(lr){2-4} \cmidrule(lr){5-6}
                & \scriptsize Task 1 & \scriptsize Task 2 & \scriptsize Task 3 & AUC & LCA \\
    		\midrule
    		Rand\textsuperscript{‡} & 99.82 & 97.92 & 92.39 & 96.71\tiny$\pm$0.00 & 95.12\tiny$\pm$0.08 \\
                FedAvg\textsuperscript{†} & 99.82 & 97.90 & 93.65 & 97.12\tiny$\pm$0.12 & 95.32\tiny$\pm$0.18 \\
                FedProx\textsuperscript{†} & 99.74 & 97.67 & 93.28 & 96.90\tiny$\pm$0.20 & 95.31\tiny$\pm$0.16 \\
                FFA-LoRA\textsuperscript{†} & 99.83 & 97.93 & \underline{94.30} & \underline{97.35}\tiny$\pm$0.04 & 95.67\tiny$\pm$0.02 \\
                FedCurv\textsuperscript{‡} & 99.82 & 97.92 & 93.30 & 97.01\tiny$\pm$0.00 & 94.96\tiny$\pm$0.10 \\
                FLwF\textsuperscript{‡} & 99.82 & \underline{98.26} & 93.18 & 97.08\tiny$\pm$0.00 & 95.07\tiny$\pm$0.01 \\
                ELCFS\textsuperscript{*} & 99.77 & 97.68 & 93.51 & 96.99\tiny$\pm$0.40 & 94.34\tiny$\pm$0.72 \\
                FedGA\textsuperscript{†} & 99.82 & 97.91 & 93.45 & 97.06\tiny$\pm$0.10 & 95.20\tiny$\pm$0.04 \\
                Per-FedAvg\textsuperscript{†} & 99.83 & 98.03 & 93.70 &  97.19\tiny$\pm$0.09 & \underline{95.91}\tiny$\pm$0.01 \\
                FedAMP\textsuperscript{†} & 99.82 & 97.87 & 93.49 &  97.06\tiny$\pm$0.09 & 95.29\tiny$\pm$0.05 \\
                APPLE\textsuperscript{‡} & 99.82 & 98.02 & 93.47 &  97.10\tiny$\pm$0.02 & 95.56\tiny$\pm$0.08 \\
                \midrule
                FedKEI & 99.82 & \textbf{98.74} & \textbf{95.62} & \textbf{98.06}\tiny$\pm$0.02 & \textbf{97.11}\tiny$\pm$0.14 \\
    		\bottomrule
  		\end{tabular}
\label{tbl:baseline_oct}
\end{table}

For the CXR-FL dataset (Table \ref{tbl:baseline_cxr}), FedKEI outperforms the strongest baselines (APPLE, Per-FedAvg) by 0.98\% and 1.16\% in AUC and LCA, respectively. All FL methods (except ELCFS) outperform Rand, highlighting the value of leveraging previous tasks (through model aggregation, knowledge preservation or personalization) in facilitating new task learning for this dataset. Our FedKEI further boosts performance by effectively learning the bi-level aggregation. Similar trends are observed on the OCT dataset (Table \ref{tbl:baseline_oct}), where our FedKEI again achieves the best performance, surpassing the strongest baselines (FFA-LoRA, Per-FedAvg) by 0.71\% and 1.20\% in AUC and LCA, respectively.\

Figure \ref{fig:overall_plot} shows the learning curves of FedKEI and the baselines across all three datasets. FedKEI consistently outperforms all baselines on each task and also enables faster task learning as observed from the learning curves. For instance, for the last task of Derm-FL (as shown in Figure 1a), our FedKEI is able to achieve the final-epoch performance of the baselines within only a few initial epochs, which means that only a fraction of the training time is required to achieve similar performance as the baselines. More concretely, the strongest baseline APPLE achieves 75.83 AUC for the last task of Derm-FL at epoch 30 in 146.9 seconds, while our FedKEI surpasses this with 76.21 AUC at epoch 4, requiring just 20.3 seconds. This advantage is attributed to the high-quality initializations generated by FedKEI (as shown by its strong performance at the start of each task), which facilitate not just better final performance but also faster adaptation to new tasks.\

\subsection{Computation \& Communication Costs}
In Table~\ref{tbl:comp_comm_cost}, we summarize the computation and communication costs on Derm-FL dataset. The computation cost is measured in two ways: (1) the average GPU execution time per task, recorded on a single NVIDIA RTX 3090 GPU with 24GB memory; and (2) the number of floating-point operations required per task, including both client-side and server-side computations, reported in teraFLOPs (TFLOPs). For communication cost, we report the theoretical download and upload size per client, where $|\theta|$ denotes the parameter size of the adapter and the head, $|\mathcal{D}_{\mathcal{T}}|$ denotes the total storage size of the task dataset, $N$ denotes the number of clients, and $K$ is the number of clusters in our method.\

From the results, we observe that FedKEI’s computation cost is comparable to the baselines, ranking around the middle among the 11 methods in terms of both GPU time and FLOPs. Compared to FedAvg, it adds only 2.92 minutes of GPU time and 10.08 TFLOPs per task. FFA-LoRA incurs the lowest TFLOPs due to partial freezing of LoRA. While the cost of FedKEI is higher than that of FedProx and FFA-LoRA, it is considerably lower than more sophisticated methods such as FLwF, FedGA, and Per-FedAvg, which involve additional inference or complex meta-learning procedures that introduce substantial computation during the task learning phase. In contrast, our FedKEI performs clustering and bi-level aggregation learning only once prior to task learning, keeping the overall computation cost relatively modest.\

For communication cost, FedKEI requires transmitting data equivalent to $K$ cluster-specific modules three times between the server and clients during bi-level aggregation learning (as shown in Figure \ref{fig:fast_slow}), and uploading the learned task-specific modules once to the server after local task learning, leading to a total communication size of $(3K+1)\times|\theta|$.  Although it is higher than methods like FedAvg involving only constant multiples of $|\theta|$, the number of clusters $K$ is typically smaller than the number of clients $N$, making FedKEI more efficient than FedCurv and APPLE. ELCFS incurs far higher cost than the others by transmitting amplitude spectra as large as the raw dataset $|\mathcal{D}_{\mathcal{T}}|$. Since $|\theta|$ consists only of lightweight components (e.g., a LoRA adapter on ViT-B/16 is only 0.28 MB), the actual communication cost of our FedKEI remains low—approximately 2.86 MB with $K=3$ in our experiments. \

\setlength{\tabcolsep}{4pt}
\begin{table}[t]
\centering
\caption{Computation and communication costs on Derm-FL dataset.}
  		\begin{tabular}{l c c c c c c c}
    		\toprule
\multirow{2}[2]{*}{\textbf{Method}} & \multicolumn{2}{c}{Computation Cost} & \multicolumn{1}{c}{Communication Cost}\\
\cmidrule(lr){2-3} \cmidrule(lr){4-4}
& GPU time / task & TFLOPS / task & Theoretical cost\\
\midrule
                FedAvg & 20.61 min & 128.54 & $2\times|\theta|$ \\
                FedProx & 20.66 min & 128.54 & $2\times|\theta|$ \\
                FFA-LoRA & 20.59 min & 85.68 & $|\theta|$ \\
                FedCurv & 21.50 min & 156.48 & $(2N+2)\times|\theta|$ \\
                FLwF  & 29.75 min & 192.90 & $2\times|\theta|$ \\
                ELCFS  & 36.23 min & 130.68 & $(N+1)\times|\mathcal{D}_{\mathcal{T}}|$ \\
                FedGA  & 32.39 min & 157.10 & $2\times|\theta|$ \\
                Per-FedAvg  & 36.64 min & 170.48 & $4\times|\theta|$ \\
                FedAMP  & 20.92 min & 128.78 & $2\times|\theta|$ \\
                APPLE  & 21.39 min & 149.64 & $(N+1)\times|\theta|$ \\
                \midrule
                FedKEI  & 23.52 min & 138.62 & $(3K+1)\times|\theta|$ \\
    		\bottomrule
  		\end{tabular}
\label{tbl:comp_comm_cost}
\end{table}

\subsection{Ablation Studies}
\setlength{\tabcolsep}{2.5pt}
\begin{table*}[t]
\centering
\caption{Ablation studies of FedKEI's key components. Each variant includes one more design at a time. For ease of comparison, we also report the \textbf{performance increment of each variant over the previous version} after adding a new component.}
\begin{tabular}{l c c c c c c c c c c c c c c c c}
\toprule
\multirow{2}[2]{*}{\textbf{Variant}} & \multirowcell{2}{learned \\ $\boldsymbol{\alpha}$} & \multirowcell{2}{modules\\ clustering} & \multirowcell{2}{learned \\ $\boldsymbol{\beta}$} & \multirowcell{2}{bi-level\\ optimization} & \multicolumn{4}{c}{\textbf{Derm-FL}} & \multicolumn{4}{c}{\textbf{CXR-FL}} & \multicolumn{4}{c}{\textbf{OCT}}\\
\cmidrule(lr){6-9} \cmidrule(lr){10-13} \cmidrule(lr){14-17}
& & & & & AUC & Incr. & LCA & Incr. & AUC & Incr. & LCA & Incr. & AUC & Incr. & LCA & Incr.\\
\midrule
Rand & & & & & 83.57\tiny$\pm$0.12 & - & 81.15\tiny$\pm$0.05 & - & 81.17\tiny$\pm$0.44 & - & 78.40\tiny$\pm$0.54 & - & 96.71\tiny$\pm$0.00 & - & 95.12\tiny$\pm$0.08 & - \\
\midrule
Variant A & \checkmark & & & & 85.62\tiny$\pm$0.00 & 2.05 & 84.19\tiny$\pm$0.01 & 3.04 & 81.97\tiny$\pm$0.23 & 0.8 & 80.15\tiny$\pm$0.09 & 1.75 & 96.93\tiny$\pm$0.20 & 0.22 & 95.84\tiny$\pm$0.30 & 0.72 \\
Variant B & \checkmark & \checkmark & & & 85.97\tiny$\pm$0.03 & 0.35 & 84.24\tiny$\pm$0.08 & 0.05 & 82.11\tiny$\pm$0.40 & 0.14 & 80.28\tiny$\pm$0.32 & 0.13 &  97.37\tiny$\pm$0.32 & 0.44 & 96.33\tiny$\pm$0.40 & 0.49\\
Variant C & \checkmark & \checkmark & \checkmark & & 86.02\tiny$\pm$0.12 & 0.05 & 84.41\tiny$\pm$0.12 & 0.17 & 82.49\tiny$\pm$0.07 & 0.38 & 80.61\tiny$\pm$0.13 & 0.33 & 97.50\tiny$\pm$0.23 & 0.13 & 96.43\tiny$\pm$0.09 & 0.10\\
\midrule
FedKEI & \checkmark & \checkmark & \checkmark & \checkmark & \textbf{86.52}\tiny$\pm$0.02 & 0.50 & \textbf{85.07}\tiny$\pm$0.03 & 0.66 & \textbf{83.07}\tiny$\pm$0.05 & 0.58 & \textbf{81.33}\tiny$\pm$0.06 & 0.72 &  \textbf{98.06}\tiny$\pm$0.02 & 0.56 & \textbf{97.11}\tiny$\pm$0.14 & 0.68 \\
\bottomrule
\end{tabular}
\label{tbl:key_component}
\end{table*}

\subsubsection{Effect of Key Components in FedKEI}
To assess the contribution of each component in FedKEI, we compare it with three incremental variants: Variant A aggregates all previous task-specific modules within each client into client-specific modules and learns the weights $\boldsymbol{\alpha}$ to combine them for initialization; Variant B adds global clustering, replacing client-specific with cluster-specific modules for aggregation; Variant C learns the intra-cluster weights $\boldsymbol{\beta}$ via direct gradient descent on new task objectives, similarly to $\boldsymbol{\alpha}$. Our FedKEI further employs a bi-level optimization scheme, optimizing the intra-cluster weight $\boldsymbol{\beta}$ collaboratively across clients to facilitate the local adaptation of the inter-cluster weight $\boldsymbol{\alpha}$.\

Table \ref{tbl:key_component} presents ablation results on the three datasets, along with the performance gain of each variant over its predecessor. We observe consistent improvements with each added component across all datasets. Notably, learning the inter-cluster (or inter-client) weight $\boldsymbol{\alpha}$ yields significant gains over Rand, especially for Derm-FL and CXR-FL. This is because the mechanism of learning to combine past modules towards new task objectives serves to generate a much more informative initialization than the random initialization. Adding clustering further boosts performance, as the task-specific modules within each cluster are now more relevant, producing diverse clusters that enable more flexible personalized aggregation. Allowing the intra-cluster weight $\boldsymbol{\beta}$ to be learned towards new tasks further increases the flexibility of customizing the initializations. Our FedKEI, by employing the bi-level optimization scheme, shows notable improvements over Variant C, demonstrating its effectiveness in learning better bi-level weights and producing better initializations for improved task adaptation.\

\subsubsection{Effect of FM and Adapter Choices}
\setlength{\tabcolsep}{6pt}
\begin{table}[t]
\centering
\caption{Baseline comparisons on Derm-FL dataset with (1) fine-tuning \textbf{Swin-B} with LoRA, and (2) fine-tuning ViT-B/16 with \textbf{IA3}.}
  		\begin{tabular}{l c c c c}
    		\toprule
    		\multirow{2}[2]{*}{\textbf{Method}} & \multicolumn{2}{c}{\textbf{Swin-B} + LoRA} & \multicolumn{2}{c}{ViT-B/16 + \textbf{IA3}}\\
    		\cmidrule(lr){2-3} \cmidrule(lr){4-5}
                & AUC & LCA  & AUC & LCA \\
    		\midrule
    		Rand & 86.54\tiny$\pm$0.08 & 84.44\tiny$\pm$0.12 &   83.99\tiny$\pm$0.10 & 82.70\tiny$\pm$0.16 \\
                FedAvg & 87.85\tiny$\pm$0.10 & 85.15\tiny$\pm$0.17 &   84.39\tiny$\pm$0.14 & 82.19\tiny$\pm$0.21 \\
                FedProx & 87.79\tiny$\pm$0.21 & 85.38\tiny$\pm$0.31 &   84.53\tiny$\pm$0.13 & 82.56\tiny$\pm$0.18 \\
                FFA-LoRA & 88.00\tiny$\pm$0.11 & 85.50\tiny$\pm$0.25 &   - &  - \\
                FedCurv & 86.60\tiny$\pm$0.01 & 83.86\tiny$\pm$0.06 &   82.96\tiny$\pm$0.07 & 80.25\tiny$\pm$0.15 \\
                FLwF & 86.31\tiny$\pm$0.18 & 83.36\tiny$\pm$0.13 &   83.17\tiny$\pm$0.19 & 80.48\tiny$\pm$0.20 \\
                ELCFS & 86.47\tiny$\pm$0.09 & 82.96\tiny$\pm$0.16 &  83.78\tiny$\pm$0.18 & 80.56\tiny$\pm$0.25 \\
                FedGA & \underline{88.08}\tiny$\pm$0.08 & \underline{85.58}\tiny$\pm$0.10 &  84.63\tiny$\pm$0.11 & 82.64\tiny$\pm$0.11 \\
                Per-FedAvg & 87.96\tiny$\pm$0.10 & 85.07\tiny$\pm$0.20 &   84.65\tiny$\pm$0.20 & 82.55\tiny$\pm$0.14 \\
                FedAMP & 85.60\tiny$\pm$0.02 & 81.99\tiny$\pm$0.11 &   82.86\tiny$\pm$0.07 & 78.50\tiny$\pm$0.11 \\
                APPLE & 87.31\tiny$\pm$0.11 & 84.80\tiny$\pm$0.16 & \underline{84.82}\tiny$\pm$0.09 & \underline{82.81}\tiny$\pm$0.22 \\
                \midrule
                FedKEI & \textbf{89.04}\tiny$\pm$0.10 & \textbf{87.46}\tiny$\pm$0.15 & \textbf{86.09}\tiny$\pm$0.09 & \textbf{85.56}\tiny$\pm$0.11\\
    		\bottomrule
  		\end{tabular}
\label{tbl:fm_adpt}
\end{table}
To validate FedKEI’s compatibility with different FMs and adapters, we conduct two experiments: (1) fine-tuning \textbf{Swin-B} \cite{liu2021swin} (pretrained on ImageNet) with LoRA, and (2) fine-tuning ViT-B/16 with \textbf{IA3 adapter} \cite{liu2022few}. Swin-B is a variant of Swin Transformer that employs hierarchical structure and shifted windows for multi-scale feature extraction, and IA3 adapts attention layers via learned rescaling vectors for the keys and values. The results on Derm-FL are summarized in Table~\ref{tbl:fm_adpt}.\

From the results, we observe two key findings. First, Swin-B with LoRA outperforms ViT-B/16 with LoRA (Table~\ref{tbl:baseline_derm}), likely due to its superior multi-scale feature processing for medical images, while ViT-B/16 with IA3 achieves performance comparable to LoRA. Second, FedKEI consistently achieves top performance: with Swin-B + LoRA, it surpasses the second-best performer (FedGA)) by 0.96\% in AUC and 1.88\% in LCA; with ViT-B/16 + IA3, it outperforms the second-best performer (APPLE) by 1.27\% in AUC and 2.75\% in LCA. These results highlight FedKEI’s robustness and adaptability across diverse FMs and adapters. Other baseline trends remain consistent with Table~\ref{tbl:baseline_derm}. Note that since FFA-LoRA is designed specifically for LoRA, it is omitted from IA3 experiments.\

\subsubsection{Effect of Task Order}
In this section, we assess FedKEI’s robustness to task order variations on Derm-FL. The first experiment retains synchronous alignment but reverses the task order for each of the four source datasets (e.g., ISIC19: DF → BKL → AK → BCC → NV). The second adopts an asynchronous setup, randomly shuffling the task order for each client so that all clients follow different task orders.
The former tests the effect of task order in the synchronous setup, while the latter tests the asynchronous setup, where a disease encountered by one client may have been seen by other clients.\

As shown in Table~\ref{tbl:task_order}, reversing the task order slightly improves overall performance compared to the original (Table~\ref{tbl:baseline_derm}), suggesting that learning certain tasks earlier may benefit subsequent ones—likely due to shared visual features or semantic similarities that enhance knowledge transfer. In this setting, FedKEI outperforms all baselines, with margins of 1.40\% in AUC and 1.58\% in LCA over the second-best performer Per-FedAvg. Shuffling task orders across clients yields notable AUC gains of 1.5–3\% compared to Table~\ref{tbl:baseline_derm}, likely because when clients follow different task orders, a new task encountered by one client may have already been learned by others, allowing it to benefit from previously learned modules. Under this asynchronous setup, FedKEI again achieves the best results, surpassing the next-best methods (APPLE, FedGA) by 1.08\% in AUC and 1.81\% in LCA, demonstrating its robustness across different task order settings.\

\setlength{\tabcolsep}{6pt}
\begin{table}[t]
\centering
\caption{Baseline comparisons on Derm-FL dataset with (1) synchronous \textbf{reversed} task order, and (2) asynchronous \textbf{shuffled} task orders across clients.}
  		\begin{tabular}{l c c c c}
    		\toprule
    		\multirow{2}[2]{*}{\textbf{Method}} & \multicolumn{2}{c}{\textbf{Reversed} Task Order} & \multicolumn{2}{c}{\textbf{Shuffled} Task Order}\\
    		\cmidrule(lr){2-3} \cmidrule(lr){4-5}
                & AUC & LCA  & AUC & LCA \\
    		\midrule
    		Rand & 85.01\tiny$\pm$0.11 & 82.99\tiny$\pm$0.18 & 86.44\tiny$\pm$0.13 & 85.34\tiny$\pm$0.20 \\
                FedAvg & 85.24\tiny$\pm$0.08 & 83.39\tiny$\pm$0.13 &   86.57\tiny$\pm$0.21 & 85.42\tiny$\pm$0.19 \\
                FedProx & 85.14\tiny$\pm$0.22 & 83.31\tiny$\pm$0.17 &   86.64\tiny$\pm$0.09 & 85.52\tiny$\pm$0.13 \\
                FFA-LoRA & 85.01\tiny$\pm$0.23 & 83.44\tiny$\pm$0.15 &   86.85\tiny$\pm$0.22 & 85.57\tiny$\pm$0.31 \\
                FedCurv & 85.07\tiny$\pm$0.09 & 82.85\tiny$\pm$0.09 &   86.39\tiny$\pm$0.12 & 84.70\tiny$\pm$0.15 \\
                FLwF & 84.91\tiny$\pm$0.16 & 82.80\tiny$\pm$0.20 &  86.46\tiny$\pm$0.11 & 84.67\tiny$\pm$0.19 \\
                ELCFS & 84.36\tiny$\pm$0.05 & 81.58\tiny$\pm$0.11 & 85.87\tiny$\pm$0.24 & 82.62\tiny$\pm$0.21 \\
                FedGA & 85.28\tiny$\pm$0.19 & 83.41\tiny$\pm$0.11 & 86.79\tiny$\pm$0.07 & \underline{85.76}\tiny$\pm$0.13 \\
                Per-FedAvg & \underline{85.38}\tiny$\pm$0.08 & \underline{83.76}\tiny$\pm$0.15 &  86.66\tiny$\pm$0.06 & 85.38\tiny$\pm$0.14\\
                FedAMP & 84.40\tiny$\pm$0.01 & 81.92\tiny$\pm$0.04 &   86.29\tiny$\pm$0.04 & 84.63\tiny$\pm$0.05 \\
                APPLE & 85.37\tiny$\pm$0.06 & 83.62\tiny$\pm$0.18 & \underline{86.86}\tiny$\pm$0.11 & 85.74\tiny$\pm$0.15 \\
                \midrule
                FedKEI & \textbf{86.78}\tiny$\pm$0.07 & \textbf{85.34}\tiny$\pm$0.11 &  \textbf{87.94}\tiny$\pm$0.14 & \textbf{87.57}\tiny$\pm$0.09  \\
    		\bottomrule
  		\end{tabular}
\label{tbl:task_order}
\end{table}

\subsection{Qualitative Analysis}
In this section, we qualitatively analyze FedKEI's capability of learning meaningful initializations by visualizing the clustering results and the learned inter-cluster weights $\boldsymbol{\alpha}$ for assigning importance to different clusters. For this analysis, we use CXR-FL dataset, which consists of three clients and six tasks for each client. We investigate the quality of the inter-cluster weight $\boldsymbol{\alpha}$ learned for the last task (i.e., the 6$^{th}$ task) by assessing how closely $\boldsymbol{\alpha}$ aligns with the optimized adapter and head of the 6$^{th}$ task (obtained after local fine-tuning) in terms of assigning importance to different clusters to generate the initializations. Here, we set the number of clusters $K=3$ for both adapters and heads.\

\begin{figure}[t]
\centerline{\includegraphics[width=0.95\linewidth]{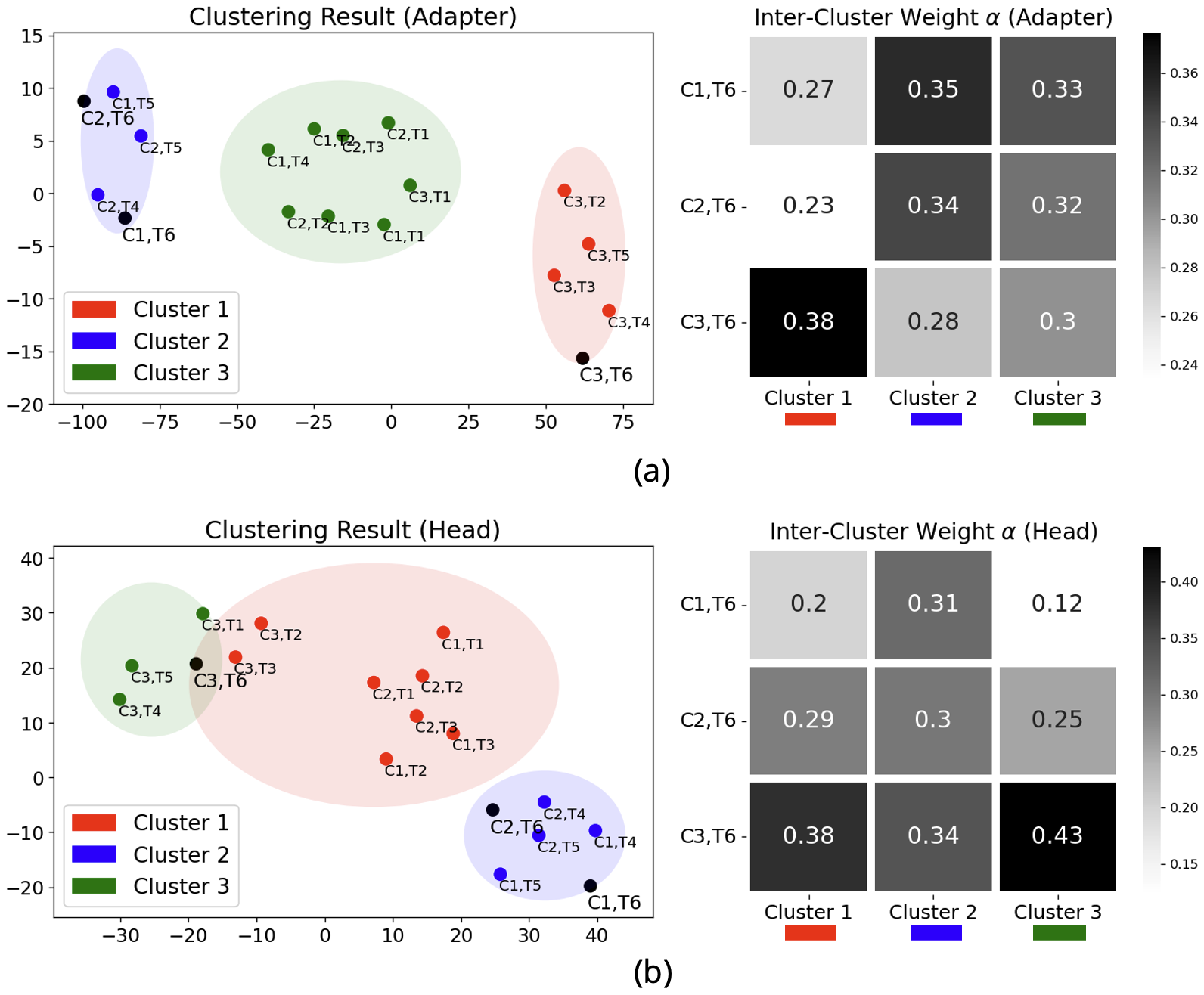}}
\caption{Clustering results (t-SNE) and the inter-cluster weight $\boldsymbol{\alpha}$ learned for the 6$^{th}$ task of three clients of CXR-FL for (a) adapter and (b) head. Each point (C$i$,T$t$) is the task-specific module of task $t$ of client $i$.}
\label{fig:clus_plot}
\end{figure}

Figure \ref{fig:clus_plot} shows the t-SNE plots of the clustering results and the learned $\boldsymbol{\alpha}$ for the 6$^{th}$ task of three clients for the adapters and heads, respectively. In the t-SNE plots, each point (C$i$,T$t$) represents the task-specific module of task $t$ of client $i$, obtained after local fine-tuning on that task. To generate initializations for the 6$^{th}$ task, we perform clustering on all the $3\times 5 = 15$ previous task-specific modules stored at the server. The clustering results are shown in three different colors, and the black dots represent the optimized modules of the 6$^{th}$ task of three clients after local fine-tuning (which are not involved in the clustering). From the plots, we see that FedKEI tends to assign larger weights to clusters closer to the optimized modules of the 6$^{th}$ task. For instance, in Figure \ref{fig:clus_plot}a, the optimized adapter of the 6$^{th}$ task of client 3 (i.e., point C3,T6) is located closer to cluster 1 and more distant from cluster 2. The weight $\boldsymbol{\alpha}$ learned to generate the adapter's initialization for this task assigns a higher value to cluster 1 and a lower value to cluster 2. Similar trends are also observed in Figure \ref{fig:clus_plot}b for the heads. This alignment shows that FedKEI effectively identifies clusters more relevant to the optimally learned adapter and head of the new task, generating well-informed initializations that facilitate new task learning.\

\subsection{Evaluation on Medical Pretrained FMs and 3D Images}

To further assess FedKEI’s effectiveness, we conduct experiments under two additional setups: (1) using a larger medically-pretrained FM and (2) applying it to 3D medical imaging. For the first, we evaluate FedKEI with RETFound \cite{zhou2023foundation}, fine-tuning its ViT-L/16 encoder with LoRA on our federated OCT dataset. For the second, we use the CC-CCII dataset \cite{zhang2020clinically}, which contains CT scans from three classes: Normal, Common Pneumonia (CP), and Novel Coronavirus Pneumonia (NCP). We simulate the FCL setting by distributing each class across 5 clients using $Dir(0.5)$ and defining two sequential tasks: identifying CP from Normal, and NCP from \{Normal, CP\}. We adopt the VoCo model \cite{wu2024voco}
pretrained on 3D CT data and fine-tune it with LoRA on CC-CCII.\

\setlength{\tabcolsep}{4pt}
\begin{table}[t]
\centering
\caption{Baseline comparisons by fine-tuning LoRA on (1) RETFound with the OCT dataset, and (2) VoCo with the CC-CCII dataset.}
  		\begin{tabular}{l c c c c}
    		\toprule
    		\multirow{2}[2]{*}{\textbf{Method}} & \multicolumn{2}{c}{RETFound + LoRA on OCT} & \multicolumn{2}{c}{VoCo + LoRA on CC-CCII}\\
    		\cmidrule(lr){2-3} \cmidrule(lr){4-5}
                & AUC & LCA  & AUC & LCA \\
    		\midrule
    		Rand & 97.92\tiny$\pm$0.03 & 95.21\tiny$\pm$0.09 & 85.70\tiny$\pm$0.09 &  86.32\tiny$\pm$0.04 \\
                FedAvg & 98.85\tiny$\pm$0.08 & 96.19\tiny$\pm$0.05 & 86.09\tiny$\pm$0.05 &  87.30\tiny$\pm$0.21 \\
                FedProx & 98.66\tiny$\pm$0.07 & 96.54\tiny$\pm$0.03 & 85.74\tiny$\pm$0.14 &  86.79\tiny$\pm$0.07 \\
                FFA-LoRA & 98.77\tiny$\pm$0.13 & 96.14\tiny$\pm$0.10 & 86.19\tiny$\pm$0.08 &  87.12\tiny$\pm$0.05 \\
                FedCurv & 98.51\tiny$\pm$0.04 & 95.14\tiny$\pm$0.01 & 85.61\tiny$\pm$0.05 &  86.57\tiny$\pm$0.07 \\
                FLwF & 98.55\tiny$\pm$0.07 & 96.63\tiny$\pm$0.11 & 85.62\tiny$\pm$0.10 &  86.61\tiny$\pm$0.19 \\
                ELCFS & 98.10\tiny$\pm$0.13 & 95.52\tiny$\pm$0.04 & 85.59\tiny$\pm$0.05 &  85.74\tiny$\pm$0.17 \\
                FedGA & 98.78\tiny$\pm$0.05 & 96.20\tiny$\pm$0.09 & 86.26\tiny$\pm$0.11 &  87.31\tiny$\pm$0.10 \\
                Per-FedAvg & 98.58\tiny$\pm$0.11 & \underline{96.92}\tiny$\pm$0.05 & 86.24\tiny$\pm$0.09 &  87.52\tiny$\pm$0.05 \\
                FedAMP & 98.85\tiny$\pm$0.05 & 96.40\tiny$\pm$0.08 & 85.77\tiny$\pm$0.03 &  86.42\tiny$\pm$0.10 \\
                APPLE & \underline{98.86}\tiny$\pm$0.08 & 96.77\tiny$\pm$0.10 & \underline{86.30}\tiny$\pm$0.07 &  \underline{87.62}\tiny$\pm$0.13 \\
                \midrule
                FedKEI & \textbf{99.07}\tiny$\pm$0.07 & \textbf{96.98}\tiny$\pm$0.04 &  \textbf{87.14}\tiny$\pm$0.07 &  \textbf{88.58}\tiny$\pm$0.11 \\
    		\bottomrule
  		\end{tabular}
\label{tbl:ext}
\end{table}

The results, summarized in Table~\ref{tbl:ext}, demonstrate the robustness of FedKEI across both settings. In the first setup, FedKEI achieves top performance with RETFound, outperforming the second-best method APPLE by 0.21\% in AUC—even in a near-saturated regime where most methods approach 99\% AUC due to strong pretraining on highly relevant OCT data. In the second setup with 3D CT images, FedKEI again leads, surpassing APPLE by 0.84\% in AUC and 0.96\% in LCA. These results demonstrate FedKEI’s effectiveness with larger medical FMs and in challenging 3D imaging tasks.\

\section{Conclusion}
In this work, we propose FedKEI, a novel FL framework that enhances adaptation to new diseases for FM adapter tuning. FedKEI selectively transfers knowledge from previous task-specific modules to generate informed initializations for new tasks. It performs global clustering at the server to generalize knowledge across tasks, followed by bi-level aggregation weight learning to personalize transfer for each new task. Extensive experiments on medical datasets demonstrates FedKEI's  
advantage in adapting to new diseases compared to state-of-the-art methods.\

Despite its strong performance, FedKEI has several limitations that offer directions for future work: (1) It has yet to be evaluated in real-world large-scale FL systems, where challenges like device heterogeneity and connectivity instability may arise; (2) Testing on additional modalities and multi-modal settings (e.g., with EHRs or genomics) could further assess its generalizability; and (3) While the added overhead is moderate, future work will explore first-order approximations and model compression to improve efficiency.

\bibliographystyle{IEEEtran}
\bibliography{reference}

\end{document}